# Handling Incomplete Heterogeneous Data using a Data-Dependent Kernel

Youran Zhou, Mohamed Reda Bouadjenek, Jonathan Wells, and Sunil Aryal

Department of Science, Engineering, and Built Environment, School of Information Technology, Deakin University Geelong, Victoria, Australia

**Abstract**

Handling incomplete data in real-world applications is a critical challenge due to two key limitations of existing methods: (i) they are primarily designed for numeric data and struggle with categorical or heterogeneous/mixed datasets; (ii) they assume that data is missing completely at random, which is often not the case in practice – in reality, data is missing in patterns, leading to biased results if these patterns are not accounted for. To address these two limitations, this paper presents a novel approach to handling missing values using the Probability Mass Similarity Kernel (PMK), a data-dependent kernel, which does not make any assumptions about data types and missing mechanisms. It eliminates the need for prior knowledge or extensive pre-processing steps and instead leverages the distribution of observed data. Our method unifies the representation of diverse data types by capturing more meaningful pairwise similarities and enhancing downstream performance. We evaluated our approach across over 10 datasets with numerical-only, categorical-only, and mixed features under different missing mechanisms and rates. Across both classification and clustering tasks, our approach consistently outperformed existing techniques, demonstrating its robustness and effectiveness in managing incomplete heterogeneous data.

## 1 Introduction

Missing data is a common and critical challenge in real-world data-driven applications, arising from factors such as data collection errors, non-response in surveys, or system malfunctions. These missing values can affect all data types—numerical, categorical, or heterogeneous—complicating analysis and degrading machine learning model performance, often leading to suboptimal outcomes.

Dealing with incomplete data when designing data-driven decision-making systems is challenging, largely due to two primary limitations in current approaches. First, traditional methods for handling missing data predominantly focus on numerical data [1, 2, 3, 4] and often extend to categorical or mixed-type datasets through additional preprocessing steps and transformations (e.g., label encoding, one-hot encoding, and ordinal encoding) [5, 6]. While these transformations enable the application of numerical methods to categorical data, they frequently lead to increased data dimensionality and computational complexity, adding significant processing overhead. Second, a significant challenge in handling missing data lies in the underlying data mechanisms. Traditional methods often assume that data is *Missing Completely at Random* (MCAR). However, in many real-world situations, data may follow the *Missing at Random* (MAR) or *Missing Not at Random* (MNAR)mechanisms, where missingness is related to observed or unobserved variables. If these patterns are not properly accounted for, they can lead to biased results. [7, 8, 9]

To address the limitations mentioned above, we propose the **Incomplete-Probability Mass Similarity Kernel (PMK)**, a novel representation learning-based approach specifically designed for *incomplete heterogeneous data*. Representation learning provides a powerful alternative to traditional imputation techniques by focusing on creating robust data representations that directly support downstream tasks without explicitly reconstructing missing values. This approach not only simplifies data processing but also preserves data privacy by eliminating the need for direct imputation of missing information. In addition, PMK leverages inherent assumptions about data distributions and missing mechanisms to perform efficient similarity representation learning, unifying the representation of different data types and capturing meaningful pairwise similarities between data points. This capability enables PMK to significantly enhance performance across a variety of downstream tasks.

We evaluated PMK a diverse set of over ten datasets, including numerical-only, categorical-only, and mixed-type features, under various missing data mechanisms and different missing rates. The results demonstrate that PMK consistently outperforms traditional methods in both classification and clustering tasks, proving its robustness and effectiveness in managing incomplete heterogeneous data.

The key contributions of this work are as follows:

- **Simple and Efficient Kernel for Incomplete Data**: We propose a fast, direct kernel method that bypasses imputation, focusing on enhancing downstream performance without the need to reconstruct missing values.

- **Solution for Heterogeneous Data**: PMK addresses the limitations of existing kernel methods, effectively managing numerical, categorical, and mixed data types within a single framework.

- **Versatile Across Missing Mechanisms**: PMK performs robustly under various missing data mechanisms (MCAR, MAR, MNAR), making it applicable to a broad range of datasets and scenarios.

- **Comprehensive Evaluation Across Datasets and Tasks**: We evaluated our approach on 16 datasets with various data types and missing mechanisms, consistently outperforming existing techniques in both classification and clustering tasks.

## 2 Related Work

There is a substantial body of research related to missing data imputation. In the following, we review the major work related to various imputation techniques, including statistical methods, machine learning approaches, and kernel-based models, with a focus on their effectiveness in handling different types of incomplete data across diverse domains.

**Traditional imputations:** Handling missing data remains a significant challenge in data-driven fields, particularly when dealing with heterogeneous datasets that encompass numerical, categorical, and mixed-type data. Traditional imputation techniques, such as mean imputation, MICE, EM algorithm and k-nearest neighbors, are widely used to estimate missing values from the observed data [1, 2, 3, 4]. While these methods are straightforward to implement, they often require substantial preprocessing and fail to capture the inherent complexities present in heterogeneous datasets. Furthermore, these approaches frequently distort the original data distribution, leading to suboptimal model performance.

**Generative models for imputation:**Recent advancements in deep learning have introduced more sophisticated methods, including generative models like GAIN [10], MIWAE [11], not-MIWAE [12], and CSDI [13]. These models excel at imputing missing data in contexts such as image processing and time series. However, they are predominantly designed for numerical data and struggle to generalize to heterogeneous datasets. HIVAE [15], a variational autoencoder-based model, attempts to handle mixed-type data but relies on imputation and pre-defined distributions, which limits its adaptability in more complex applications.

Table 1: Comparison of Methods Across Key Criteria

| Criteria | PMK | [1] | [3] | [4] | [15] | [25] |
|---|---|---|---|---|---|---|
| Cat/Het | ✓ | ✓ | ✗ | ✗ | ✓ | ✗ |
| No Basic Imp | ✓ | - | - | ✓ | ✗ | ✗ |
| MAR/MNAR | ✓ | ✗ | ✗ | ✗ | ✓ | ✓ |

**Kernel methods for incomplete data:**Kernel methods, which map data into latent feature spaces to reveal hidden patterns, have also been applied to handle incomplete data. Smola et al.[19] and Pelckmans et al.[20] introduced early approaches for adapting Support Vector Machines and Gaussian Processes to missing data, using exponential families in feature space. Chechik et al.[21] proposed a kernel that adapts by focusing solely on the observed components of input vectors. An extension to the RBF kernel for handling missing data was proposed by Śmieja et al. [24, 25], which incorporates uncertainty in missing attributes through regularization. While this method improves the handling of incomplete data, it still cannot effectively manage heterogeneous data types. Additionally, existing kernel-based methods often fall short when applied to complex datasets and missing mechanism.

**Data-dependent kernels:**In contrast to traditional kernels, another types of kernel called, data-dependent kernels that measure similarity based on both distance and local data distribution [26, 27, 28, 29]. Although these kernels have demonstrated success in clustering tasks, their applicability to handling incomplete data has been limited.

**Challenges with missing mechanisms:**Another challenge in the realm of missing data is understanding the missing mechanism, which explains how and why data points become missing. According to Rubin's theory [31], there are three types of missing mechanisms: Missing Completely At Random (MCAR), Missing At Random (MAR), and Missing Not At Random (MNAR). MCAR is the simplest case, as it assumes that the missing data is completely random. In contrast, MAR and MNAR are more complex, as they involve assumptions regarding the relationship between the missing data and known information or the missing values themselves. These mechanisms can sometimes be exceedingly complex, complicating the handling of missing data. While there are approaches designed to address various missing mechanisms [11, 12], most of them are based on the assumption of numerical data and do not effectively apply to categorical or heterogeneous dataset.

Our proposed PMK is a data-dependent similar-ity kernel specifically designed to address the challenges of incomplete heterogeneous data with various missing mechanisms. As detailed in Table 1, we compare PMK against traditional imputation methods [1, 3, 4], a generative models [15], and a kernel-based method [25] using three key criteria: (1) the ability to handle categorical or heterogeneous data, (2) the capability to directly process incomplete data without relying on prior imputation or a mask matrix, and (3) the flexibility to manage various missing data mechanisms. PMK stands out as the only method that comprehensively addresses all three aspects, making it highly versatile and efficient for handling complex datasets involving different types of missing data. This adaptability highlights PMK's robustness, providing a more complete solution where other methods may fall short.

## 3 Data-Dependent Kernel

In this section, we introduce our proposed Incomplete-**P**robability **M**ass Similarity **K**ernel (PMK), a data-dependent kernel designed to handle incomplete heterogeneous data.

**3.1 Probability Mass Kernel (PMK):** The Probability Mass Kernel (PMK) builds upon the concept of $m_0$-dissimilarity [26, 30], which extends traditional distance metrics by incorporating the data distribution surrounding the objects being compared. Unlike traditional data-independent kernels, such as Gaussian or Laplacian kernels–where the same distance between two points always results in the same similarity–data-dependent kernels adjust similarity based on the density of data in the surrounding region. For example, in a sparse region of the dataset, two points at the same distance may be considered more similar than if they were in a densely populated area. This ability to adapt to the underlying data distribution allows data-dependent kernels to better capture the structure and patterns of complex datasets, resulting in more accurate similarity measures.

Let $X$ be a dataset consisting of $m$ instances, each represented in $n$ dimensions. The $i$-th instance can be denoted as a vector $\mathbf{x}^{(i)} = \langle x_1^{(i)}, x_2^{(i)}, \ldots, x_n^{(i)} \rangle$, where $i \in [1, m]$. For a given feature $k$, let $R_k(x_k^{(i)}, x_k^{(j)})$ represent the region covering instances $\mathbf{x}^{(i)}$ and $\mathbf{x}^{(j)}$ for feature $k$. Specifically, the size of this region, $|R_k(x_k^{(i)}, x_k^{(j)})|$, represents the number of data points whose $k$-th feature values lie within this range. This count quantifies local data density, enhancing the similarity measure by incorporating contextual information.

The $m_0$-dissimilarity between $\mathbf{x}^{(i)}$ and $\mathbf{x}^{(j)}$ is calculated as:

$$m_0(\mathbf{x}^{(i)}, \mathbf{x}^{(j)}) = \left( \frac{1}{n} \sum_{k=1}^{n} \log \frac{|R_k(x_k^{(i)}, x_k^{(j)})|}{m} \right) \tag{3.1}$$

Intuitively, if many data points fall within the region, the two instances are considered more dissimilar, as a dense data distribution surrounds them. Conversely, sparse regions yield small dissimilarity scores.

To efficiently compute $|R_k(x_k^{(i)}, x_k^{(j)})|$, each feature $k$ is discretised into $b$ bins, and a pre-computed bin data mass is used. A matrix stores the data masses between all bin pairs for each feature, enabling fast lookups to determine the region size and significantly speeding up the computation process.

Finally, the $m_0$-dissimilarity is normalized to ensure it meets the properties of a valid self-similarity and symmetry metric. The resulting Probability Mass-based Kernel similarity is defined as [26, 30]:

$$PMK(\mathbf{x}^{(i)}, \mathbf{x}^{(j)}) = \frac{2 \times m_0(\mathbf{x}^{(i)}, \mathbf{x}^{(j)})}{m_0(\mathbf{x}^{(i)}, \mathbf{x}^{(i)}) + m_0(\mathbf{x}^{(j)}, \mathbf{x}^{(j)})} \tag{3.2}$$

**3.2 PMK for Heterogeneous Data:** The original PMK was designed primarily for numerical data. However, to better accommodate heterogeneous data—datasets that contain a mix of numerical and categorical values—this method was further extended in [30]. In Equation 3.1, $\frac{|R_k(x_k^{(i)}, x_k^{(j)})|}{m}$ is the probability mass in the region, denoted as $P(R_k(x_k^{(i)}, x_k^{(j)}))$, which is calculated by discretizing the values of the numerical features. As suggested in [30], for categorical features, $P(R_k(x_k^{(i)}, x_k^{(j)}))$ can be computed using probabilities of categorical labels. The computation depends on whether the categorical feature $k$ is **ordinal** (where values follow a natural order, such as size = {S, M, L, XL, XXL}) or **nominal** (where values have no inherent order, such as color = {Red, Green, Blue, Yellow, White}), as outlined below:

$$P(R_k(x_k^{(i)}, x_k^{(j)})) = \begin{cases} \displaystyle\sum_{z_k = \min(x_k^{(i)}, x_k^{(j)})}^{\max(x_k^{(i)}, x_k^{(j)})} P(z_k), & \text{ordinal } k \\ P(x_k^{(i)} \vee x_k^{(j)}), & \text{nominal } k \end{cases} \tag{3.3}$$

where $P(z_k)$ represents the frequency of the feature value $z_k$ divided by the total number of instances $m$, and $P(x_k^{(i)} \vee x_k^{(j)})$ denotes the probability of a feature $k$ having the label of $x_k^{(i)}$ or $x_k^{(j)}$.

### 3.3 PMK for Incomplete Data:

In the previous section, we discussed how PMK operates in heterogeneous domains with complete data. However, dealing with incomplete data is not straightforward. To address this, we introduce the following adjustments to adapt the PMK framework for missing data.

**Record the frequency of missing data in each feature**: Since PMK is based on the probability mass, we accumulate missing data in each feature $k$ into a separate bucket, denoted as $\mathcal{B}_k$, and record its mass $|\mathcal{B}_k|$ to use later while computing similarity of data objects.

**Adjusting the Probability Mass Calculation for Incomplete Data**: Next, we propose a modification to the estimation of probability mass between $x_k^{(i)}$ and $x_k^{(j)}$, $P(R_k(x_k^{(i)}, x_k^{(j)}))$, when one or both of them have missing value in feature $k$. We take a very conservative approach and assign maximal possible dissimilarity (i.e., large probability mass) based on the available information.

*Only one of them is missing*: If $k$ is numeric or ordinal, we can treat them similarly because numerical data is converted into ordinal through discretization. $|R_k(x_k, ?)|$ can be estimated by looking at the data masses between the bin of $x_k$ ($Bin(x_k)$) and the first (minimum value) and last (maximum value) bins. Let $\mathcal{M}_L(x_k)$ and $\mathcal{M}_R(x_k)$ are data masses on the left and right sides of $Bin(x)$, including itself (as shown in Figure 1). Given that '?' could be anything, we take a conservative approach and assign maximal dissimilarity as:

$$|R_k(x_k, ?))| = \max(\mathcal{M}_L(x_k), \mathcal{M}_R(x_k)) + |\mathcal{B}_k|$$

Similarly, if $k$ is nominal, $|R_k(x_k, ?)|$ can be computed using frequencies of nominal labels as:

$$|R_k(x, ?)| = \mathcal{M}(x_k) + \max_{a \in \mathcal{S}_k} \mathcal{M}(a) + |\mathcal{B}_k|$$

where, $\mathcal{M}(x_k)$ is the frequency of label $x_k$ and $\mathcal{S}_k$ is a set of possible nominal labels for feature $k$. Here, we assume '?' could be the most frequent label; hence, they could be maximally dissimilar. In both cases, we need to include the frequency of missing values ($|\mathcal{B}_k|$) because they could be any values possible.

*Both of them are missing*: In the case of numerical or ordinal features, the missing values could be the minimum and maximum. Therefore, we assign the maximum dissimilarity between them as $|R_k(?, ?)| = m$. Note that $m$ includes the number of missing values ($|\mathcal{B}_k|$) too. Similarly, in the case of nominal features, they could be the most frequent values. So, we use the frequency of the most frequent label and the number of missing values as: $|R_k(?, ?)| = \max_{a \in \mathcal{S}_k} \mathcal{M}(a) + |\mathcal{B}_k|$.

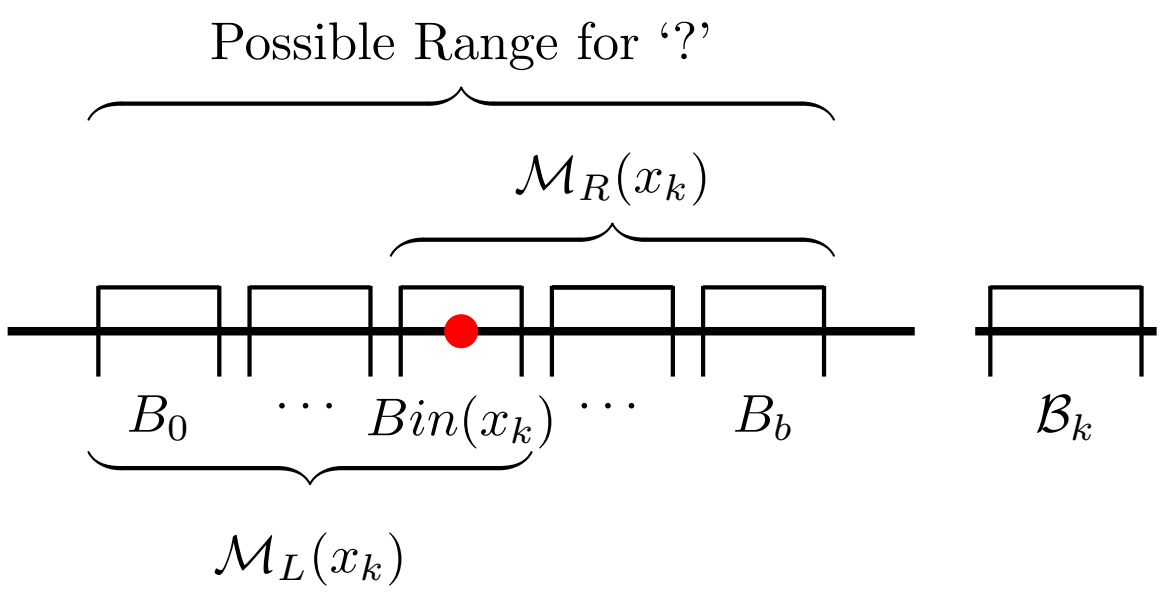

Figure 1: Adjusting the Probability Mass when a single data point is missing in a numeric or ordinal feature $k$. The red point represents the observed data point $x$, while $B_0$ and $B_b$ denote the first and last bins, respectively. Additionally, $\mathcal{B}_k$ indicates the separate bucket designated for missing data in feature $k$.

## 4 Experiments

In this section, we evaluate the performance of our proposed method on both classification and clustering tasks using 16 real-world datasets. Additional results are provided in Appendix A due to space constraints. The implementation of our method is accessible Here.

### 4.1 Experimental Setup

#### 4.1.1 Datasets:

We evaluate our model using a total of 16 datasets obtained from the UCI Machine Learning Repository[1]. For classification tasks, we utilize 10 complete datasets comprising numerical, categorical, and mixed data types. To thoroughly assess the robustness of our approach, we generated missing values across these datasets, applying different missing data mechanisms (MCAR, MAR, MNAR) and varying missing rates. This experimental setup allows us to test the model's adaptability and effectiveness in a wide range of controlled missing data scenarios.

In addition to the above datasets, we further evaluate our model on six real-world incomplete datasets to test its performance in practical settings where missing data naturally occurs. Classification tasks are performed on both complete and incomplete datasets.

Table 2 provides a detailed summary of the datasets, including their dimensions and feature types. This comprehensive evaluation highlights our model's ability to operate effectively under both simulated and real-world conditions.

#### 4.1.2 Missing Data Generation:

In our experiments, we generate missing data in the complete dataset according to three different mechanisms: Missing Completely at Random (MCAR), Missing at Random (MAR), and Missing Not at Random (MNAR). The missing rates considered are set at 5%, 10%, 20%, 30%, 40%, 50%, 60%, 70%, and 80%, allowing us to analyse the model's performance across various levels of data sparsity.

[1]https://archive.ics.uci.edu/

Table 2: Summary of Complete and Incomplete Datasets. 'N' denotes the number of instances, while 'Ord', 'Nom', and 'Num' indicate the counts of ordinal, nominal, and numerical features, respectively. 'C' represents the number of classes, and 'M.R.' signifies the missing rate.

| Complete Datasets | | | | | | |
|---|---|---|---|---|---|---|
| Dataset | N | Ord | Nom | Num | C | M.R. |
| Adult | 48,842 | 1 | 7 | 6 | 4 | - |
| Aust | 69,014 | 0 | 8 | 6 | 2 | - |
| Bank | 1,372 | 0 | 0 | 5 | 2 | - |
| Breast | 2,869 | 4 | 5 | 0 | 2 | - |
| Car | 1,728 | 6 | 0 | 0 | 4 | - |
| Heart | 303 | 0 | 8 | 5 | 5 | - |
| Sonar | 208 | 0 | 0 | 60 | 2 | - |
| Spam | 4,601 | 0 | 0 | 57 | 2 | - |
| Student | 649 | 11 | 16 | 2 | 5 | - |
| Wine | 4,898 | 0 | 0 | 12 | 2 | - |
| Incomplete Datasets | | | | | | |
| Dataset | N | Ord | Nom | Num | C | M.R. |
| Hepat | 155 | 0 | 13 | 6 | 2 | 5.67% |
| Horse | 368 | 13 | 1 | 8 | 2 | 23.80% |
| Kidney | 400 | 2 | 10 | 12 | 2 | 10.54% |
| Mammo | 961 | 4 | 0 | 1 | 2 | 3.37% |
| Pima | 768 | 0 | 0 | 8 | 2 | 12.24% |
| Wincon | 699 | 9 | 0 | 0 | 2 | 0.25% |

**MCAR and MAR Generation:** For MCAR, data is randomly removed from the dataset, ensuring that the missing values are uniformly distributed. This simulates a scenario where the absence of data is completely independent of both the values themselves and their relationships with other features. In contrast, for MAR, missing data is introduced based on the values of other features, utilizing the methodology from [34].

**MNAR Generation:** Generating MNAR data is more complex, as missing values depend on the unobserved values themselves. We implement a column-wise MNAR generation strategy, where data is missed based on its value. The MNAR generation process varies slightly according to the feature type:

- **Numerical features**: We calculate a percentile threshold for each column and selectively remove values above or below this threshold. This simulates scenarios where certain ranges of values, particularly extreme values (either very high or very low), are more prone to being missing. This reflects real-world situations where measurement errors or data noise contribute to missing data within specific ranges.
- **Ordinal features**: Each category has a chance of being missing, with a higher likelihood for the extreme categories (i.e., the highest and lowest values).
- **Nominal features**: All categories may experience potential missingness, but one selected category is assigned a higher probability of being missing. If this category alone does not achieve the desired missing rate, additional values from other categories are randomly removed to ensure the overall target missing rate is met. This approach ensures that missingness is distributed across the dataset while maintaining a focus on a specific category.

**4.1.3 Baseline Comparison Methods:** To evaluate the performance of our proposed model, we compare it against several established methods, encompassing both imputation and kernel-based similarity techniques.

For imputation, we start with the **Mean/Mode** Imputer, a simple baseline method. We also include more advanced approaches such as Multiple Imputation by Chained Equations (**MICE**)[3] and the Expectation-Maximization (**EM**) algorithm[4]. MICE is particularly notable for its robust iterative process in filling missing data gaps, while EM focuses on maximizing the likelihood of the observed data.

For kernel-based methods, we assess the generalized RBF kernel (**genRBF**)[25], commonly utilized in scenarios involving missing data. Additionally, we evaluate Kernel PCA (**KPCA**)[33] and Probabilistic PCA (**PPCA**) [32], both of which reduce dimensionality but require prior imputation (using MICE) for application on incomplete datasets. These comparisons establish a benchmark for how well our model performs relative to traditional kernel methods adapted for handling missing data.

We did not include generative models, such as HIVAE [15], in our comparisons due to their requirement for extensive preprocessing steps to handle heterogeneous data. Other generative models [11, 12] also necessitate an imputer to create a complete data structure and generate a mass matrix indicating the locations of missing values. This added complexity detracts from the focus of our evaluation on methods that can directly manage incomplete data.

For clustering tasks, we compare our model with similarity-based methods such as Simple Similarity (**Simp**) and Gower's Similarity (**Gow**) [30]. These baseline methods are frequently used for mixed data types and provide a reference point for evaluating our model's effectiveness in real-world incomplete data scenarios.

**4.1.4 Evaluation Metrics:** In our classification tasks, we utilized Support Vector Machines (SVM) with an RBF kernel for complete datasets, while for incomplete datasets—comprised entirely of binary classes—we opted for SVM with a linear kernel. The linear kernel was selected due to the relatively small and straightforward nature of the datasets. For methods such as genRBF and PMK that depend on a precomputed similarity matrix, we employed a precomputed SVM to ensure consistency across evaluations.

To ensure robustness, we applied 5-fold cross-validation, using Accuracy and F1 Score as our primary evaluation metrics. We optimized hyperparameters, including the C value and kernel settings, through an inner 5-fold cross-validation on the training set. The best-performing configuration was then applied to the entire training set, and final model performance was assessed on the unseen test set to provide an unbiased measure of generalization.

For clustering tasks, we utilized K-means clustering, with the number of clusters K set to correspond to the number of classes in the dataset (in this case, the incomplete data are all binary, therefore, K=2). We assessed performance using Normalized Mutual Information (NMI) and Adjusted Rand Index (ARI), where ARI accounts for random clustering agreements. These metrics were chosen to evaluate how well the clustering outcomes align with the true data structure, especially in the context of missing data. Each experiment was conducted over five independent runs, and the results were averaged to reduce variability in clustering performance.

## 4.2 Analysis of Results

**Classification:** Figures 2, 3, and 4 illustrate the F1 Scores for the four most representative datasets under various missing rates and mechanisms (MCAR, MAR, and MNAR). Due to space limitations, the full data tables and additional figures are provided in the appendix. In Figure 2, the PMK model generally outperforms others, particularly on mixed-type datasets like *Adult*. On purely numerical datasets like *Wine*, PMK also shows better performance, highlighting its also effectiveness on numerical dataset. Similar trends are observed in Figures 3 and 4, where PMK continues to perform strongly across most datasets.

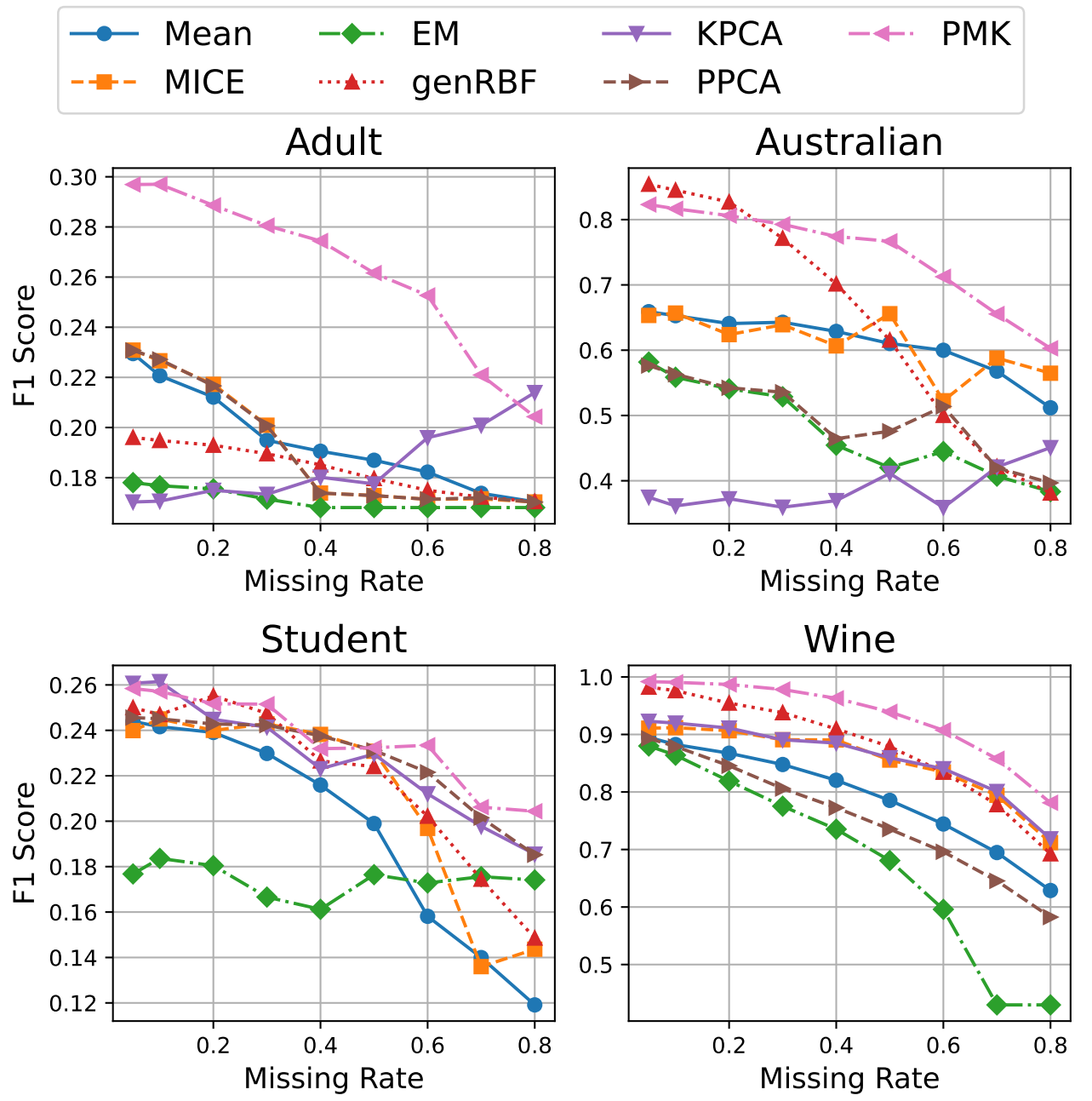

Figure 2: F1 Scores across various missing rates for four datasets (*Adult*, *Heart*, *Wine*, and *Spam*) under MCAR conditions.

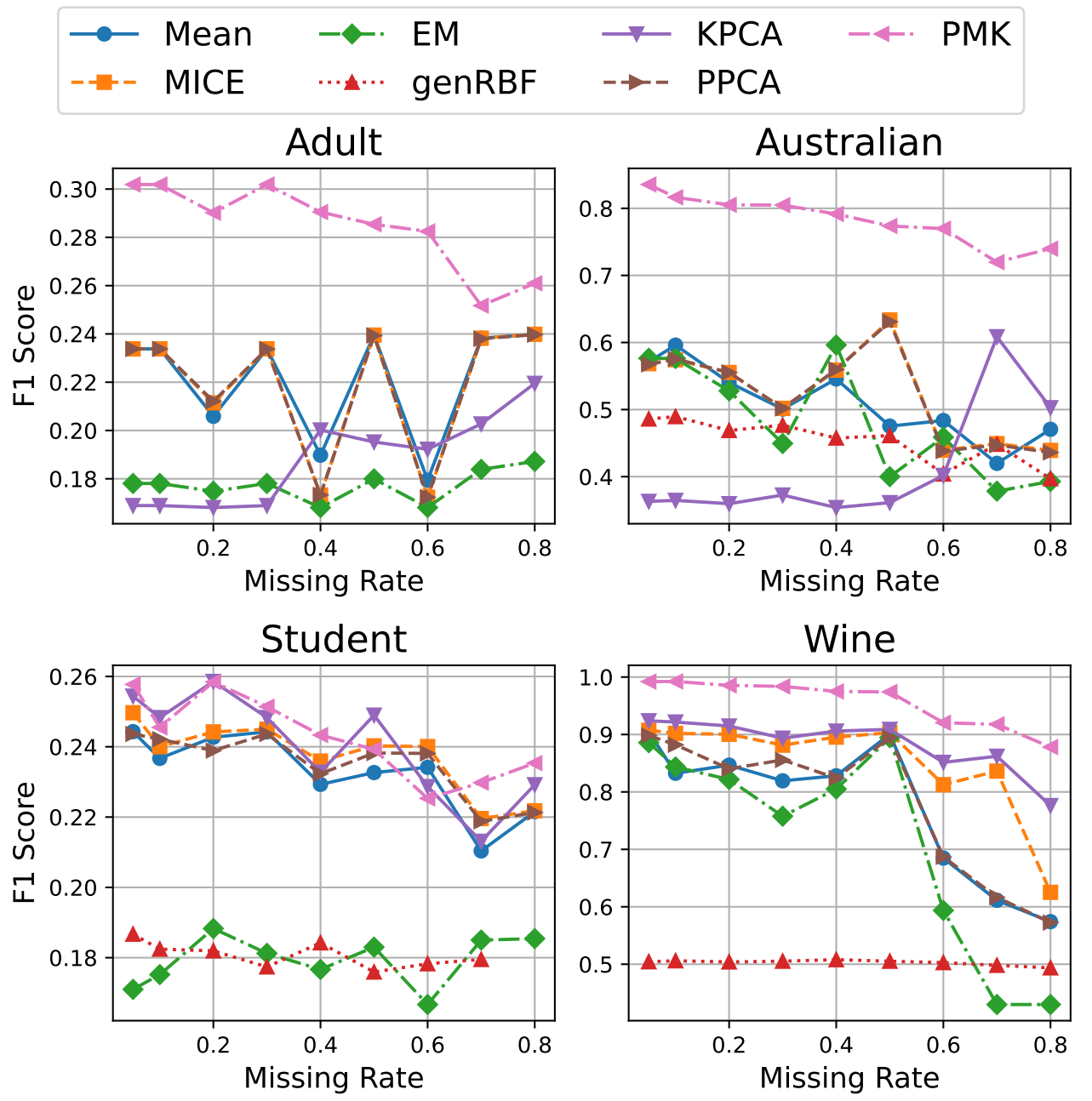

Figure 3: F1 Scores across various missing rates for four datasets (*Adult*, *Heart*, *Wine*, and *Spam*) under MAR conditions.

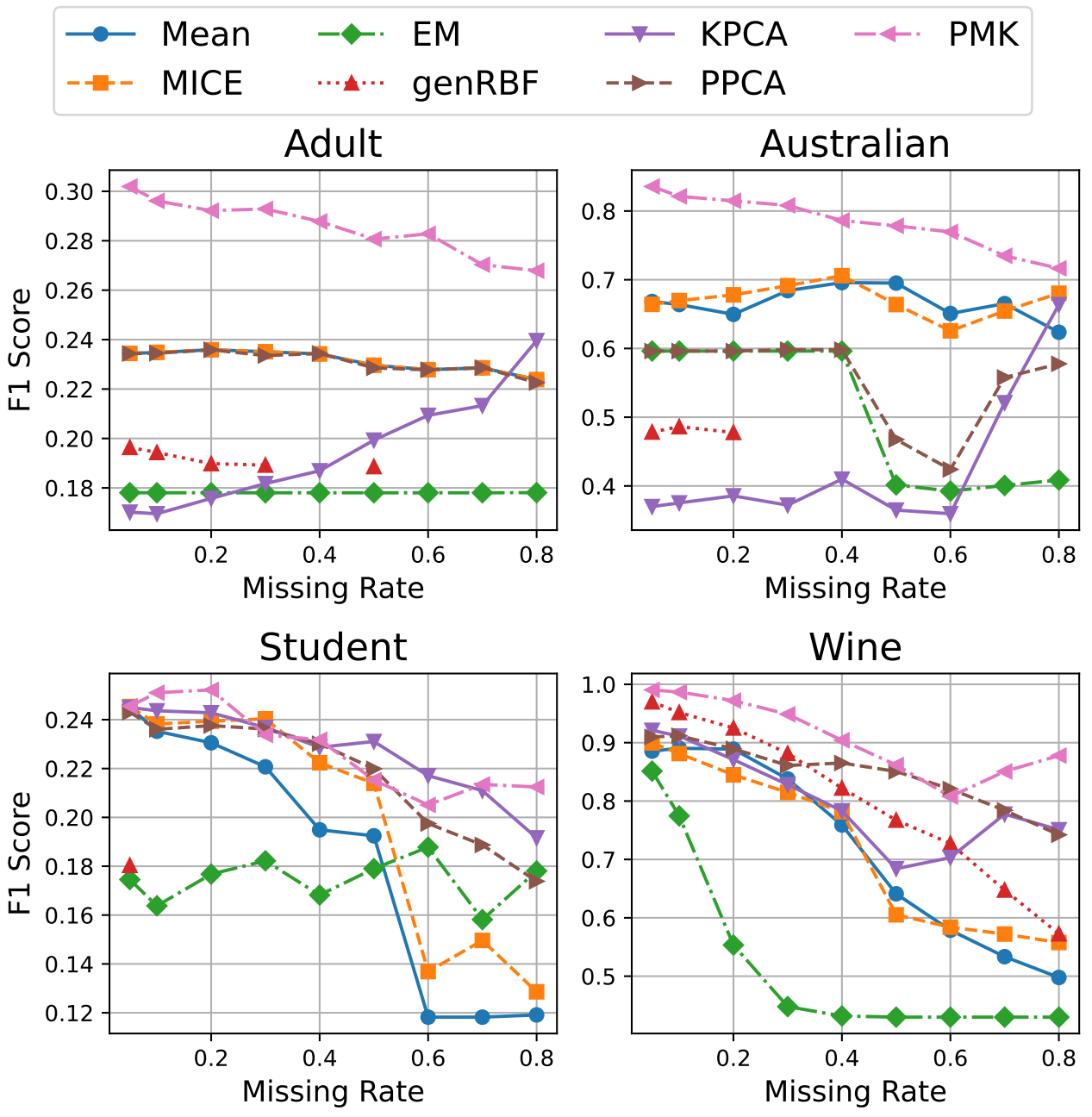

Figure 4: F1 Scores across various missing rates for four datasets (*Adult*, *Heart*, *Wine*, and *Spam*) under MNAR conditions.

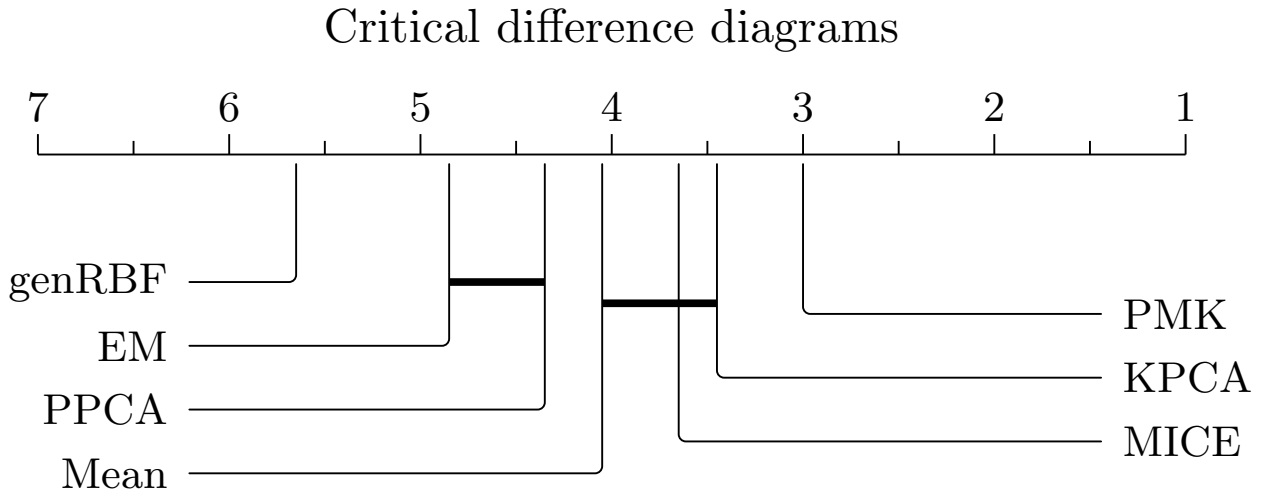

Figure 5: Critical Difference (CD) diagram showing the ranking of models based on their F1 scores across all datasets. Lower ranks represent better performance.

Table 5 presents the results at a 20% missing rate, comparing the performance of various models across different datasets and scenarios. The results obtained show that PMK generally performs well in both MCAR and MNAR cases. Although MAR does not consistently lead to the best results, it still provides acceptable performance. However, for the *Car* dataset, PMK's performance is less consistent and falls short of expectations. To provide a clearer comparison, Figure 5 employs a Critical Difference (CD) diagram to rank the models based on their F1 Scores across all datasets. The x-axis represents the rankings, with lower values indicating better performance. The PMK model consistently ranks first, followed closely by KPCA and MICE.

Table 3: Classification accuracy for incomplete datasets. Bold indicates the best result and underline indicates the second best.

| Model | Hepa | Hor | Kid | Mam | Pima | Wisc |
|---|---|---|---|---|---|---|
| PMK | 0.8065 | **0.8506** | **0.9875** | **0.8241** | 0.7618 | **0.9700** |
| Mean | <u>0.8129</u> | 0.8398 | <u>0.9775</u> | 0.8148 | **0.7735** | <u>0.9628</u> |
| MICE | <u>0.8129</u> | **0.8506** | 0.9700 | **0.8241** | <u>0.7722</u> | 0.9614 |
| EM | **0.8387** | <u>0.8424</u> | 0.9375 | <u>0.8210</u> | 0.7630 | 0.9643 |
| RBF | 0.7935 | 0.6304 | 0.6250 | 0.5140 | 0.6510 | 0.5564 |
| KPCA | 0.7935 | 0.6850 | 0.6250 | 0.8127 | 0.6510 | 0.9500 |
| PPCA | 0.8000 | **0.8506** | 0.9625 | **0.8241** | <u>0.7722</u> | 0.9628 |

Table 4: NMI scores for clustering tasks on incomplete datasets. The bold values indicate the best-performing models, while the underlined values represent the second-best performance for each dataset.

| Model | Hepa | Hor | Kid | Mam | Pima | Wisc |
|---|---|---|---|---|---|---|
| PMK | <u>0.1377</u> | <u>0.1066</u> | **0.6164** | **0.3271** | **0.1374** | **0.7585** |
| Mean | 0.0015 | 0.0078 | 0.0067 | 0.0959 | 0.0111 | 0.7295 |
| MICE | 0.0021 | 0.0024 | 0.0074 | 0.0959 | <u>0.0910</u> | 0.7427 |
| EM | 0.0025 | 0.0079 | 0.0045 | 0.0911 | 0.0211 | 0.7387 |
| RBF | 0.0772 | 0.0166 | 0.0132 | 0.0002 | 0.0052 | 0.4505 |
| KPCA | 0.0159 | 0.0583 | 0.0056 | 0.0117 | 0.0092 | 0.5186 |
| PPCA | 0.0015 | 0.0540 | 0.0081 | 0.0959 | 0.0909 | <u>0.7427</u> |
| Simp | 0.0019 | 0.0001 | 0.0424 | 0.0951 | 0.0556 | 0.0000 |
| Gow | **0.1968** | **0.1280** | <u>0.3899</u> | 0.2970 | 0.1069 | 0.6939 |

Additionally, Table 3 presents the classification accuracy for incomplete datasets, demonstrating that the PMK also yields promising results.

**Clustering:** Table 4 presents the Normalized Mutual Information (NMI) performance of various methods applied to incomplete datasets for clustering. Overall, the PMK method consistently outperforms other approaches in most scenarios. However, the relatively low missing rates in these datasets somewhat constrain its advantages in managing more complex missing data situations, thereby diminishing the potential impact of PMK's specialized handling of incomplete information.

Despite these limitations, PMK consistently delivers strong results, demonstrating its robustness even when the assumptions about the missing data mechanism are not fully satisfied. This capability suggests that PMK remains a reliable option for clustering tasks involving incomplete data, even when the underlying missing data mechanism is not clearly defined.

Table 5: Classification results showing the F1 scores of PMK and other baseline methods at a missing rate of 20% across multiple datasets and mechanisms (MCAR, MAR, and MNAR).The highest score for each dataset has been highlighted in bold, and the second-highest score is underlined.

| Model | Adult | Australian | Banknote | Breast | Car | Heart | Sonar | Spam | Student | Wine |
|---|---|---|---|---|---|---|---|---|---|---|
| **MCAR** | | | | | | | | | | |
| Mean | 0.2121 | 0.6408 | 0.8791 | 0.4127 | 0.5831 | 0.1405 | 0.7390 | 0.7576 | 0.2391 | 0.8674 |
| MICE | 0.2172 | 0.6239 | 0.9014 | 0.4127 | 0.5836 | 0.1405 | 0.7647 | 0.7420 | 0.2400 | 0.9065 |
| EM | 0.1755 | 0.5408 | 0.8582 | 0.4127 | 0.4678 | 0.1405 | 0.7692 | 0.5943 | 0.1804 | 0.8190 |
| genRBF | 0.1929 | 0.8069 | 0.8679 | 0.4126 | 0.2059 | 0.1404 | 0.3609 | 0.8456 | 0.2512 | 0.9545 |
| KPCA | 0.1750 | 0.3724 | **0.9298** | 0.4735 | **0.5841** | 0.1405 | **0.7981** | 0.6749 | 0.2448 | 0.9111 |
| PPCA | 0.2166 | 0.5424 | 0.8957 | 0.4127 | 0.3622 | 0.1405 | 0.7487 | 0.6324 | 0.2428 | 0.8455 |
| PMK | **0.2886** | **0.8161** | 0.8885 | **0.4886** | 0.4140 | **0.3060** | 0.7940 | **0.9080** | **0.2517** | **0.9869** |
| **MAR** | | | | | | | | | | |
| Mean | 0.2059 | 0.5404 | 0.9610 | 0.4127 | **0.6482** | 0.1405 | 0.7941 | 0.6160 | 0.2427 | 0.8466 |
| MICE | 0.2115 | 0.5553 | 0.9601 | 0.4127 | **0.6482** | 0.1405 | 0.7864 | 0.6312 | 0.2442 | 0.9004 |
| EM | 0.1748 | 0.5278 | 0.9499 | 0.4127 | 0.5427 | 0.1405 | 0.7573 | 0.5708 | 0.1883 | 0.8216 |
| genRBF | - | 0.4688 | 0.5161 | 0.4127 | 0.2059 | 0.1405 | 0.3480 | 0.5052 | 0.1819 | 0.5040 |
| KPCA | 0.1681 | 0.3597 | **0.9749** | **0.5297** | 0.6403 | 0.1405 | **0.8416** | 0.6676 | **0.2584** | 0.9147 |
| PPCA | 0.2121 | 0.5553 | 0.9504 | 0.4828 | 0.4018 | 0.1405 | 0.7476 | 0.6251 | 0.2389 | 0.8402 |
| PMK | **0.2902** | **0.8252** | 0.9470 | 0.4091 | 0.4249 | **0.2353** | 0.7669 | **0.9163** | **0.2584** | **0.9854** |
| **MNAR** | | | | | | | | | | |
| Mean | 0.2359 | 0.6496 | 0.8335 | 0.4127 | 0.6948 | 0.1405 | 0.7096 | 0.7959 | 0.2305 | 0.8889 |
| MICE | 0.2359 | 0.6780 | 0.8564 | 0.4127 | 0.6948 | 0.1405 | 0.7796 | 0.7710 | 0.2394 | 0.8448 |
| EM | 0.1780 | 0.5962 | 0.8804 | 0.4127 | 0.5099 | 0.1405 | 0.7583 | 0.5190 | 0.1768 | 0.5534 |
| genRBF | 0.1898 | 0.4778 | 0.7967 | - | 0.2059 | 0.1405 | 0.3464 | 0.8674 | - | 0.9252 |
| KPCA | 0.1757 | 0.3855 | **0.9757** | 0.5096 | **0.6981** | 0.1408 | **0.7939** | 0.7770 | 0.2428 | 0.8710 |
| PPCA | 0.2359 | 0.5962 | 0.9081 | 0.4127 | 0.4475 | 0.1405 | 0.7335 | 0.5630 | 0.2376 | 0.8895 |
| PMK | **0.2922** | **0.8349** | 0.8760 | **0.5160** | 0.3483 | **0.2864** | 0.7797 | **0.9414** | **0.2522** | **0.9721** |

## 5 Conclusion

In this paper, we introduced and evaluated a novel similarity measure, the incomplete-Probability Mass Kernel (**PMK**), which employs a data-dependent kernel to effectively address missing data across various data types and mechanisms. Through extensive experiments in both classification and clustering tasks, we demonstrated that PMK consistently outperforms other imputation and kernel-based approaches, particularly on datasets with mixed types. Furthermore, PMK achieved statistically comparable rankings with existing methods, highlighting its competitiveness in managing missing data.

It is important to emphasize that PMK serves primarily as a similarity measure, providing flexibility in tasks such as clustering, where no stringent assumptions about the classifier are necessary. However, this very flexibility also constrains its direct application in classification tasks. Nevertheless, PMK should be particularly well-suited for unsupervised learning scenarios, where clustering based on similarity is a primary objective.

In conclusion, PMK offers a robust and adaptable solution for tackling incomplete data, excelling in performance across diverse datasets. Its efficacy positions PMK as a promising tool for future applications in tasks involving missing data, underscoring its relevance in the realm of data-driven research.

# A Appendix

This appendix provides supplementary performance visualizations and detailed evaluation results for various models under different missing data mechanisms (MCAR, MAR, and MNAR). Both F1 scores and accuracy metrics are presented to comprehensively assess the models' performance in classification tasks. Additionally, clustering metrics, including Normalized Mutual Information (NMI) and Adjusted Rand Index (ARI), are provided to further evaluate model effectiveness in handling incomplete data scenarios. The figures and tables below expand on the results discussed in the main text, offering a more granular view of model behavior across diverse datasets.

## Classification

Figure 6 7 8 9 10 11 shows the Accuracy and F1 score for 10 dataset under three different missing mechanism. Table 9, 10, 11, 12, 13,14, 15, F1 Scores with corresponding standard deviations across various models and missing value proportions for the different dataset. The results presented in Table 8 showcase the classification accuracy of each model under different missing value scenarios.

## Clustering

For clustering tasks, Table 6 provides the NMI performance results, while Table 7,presents the ARI performance results across different methods and incomplete datasets.

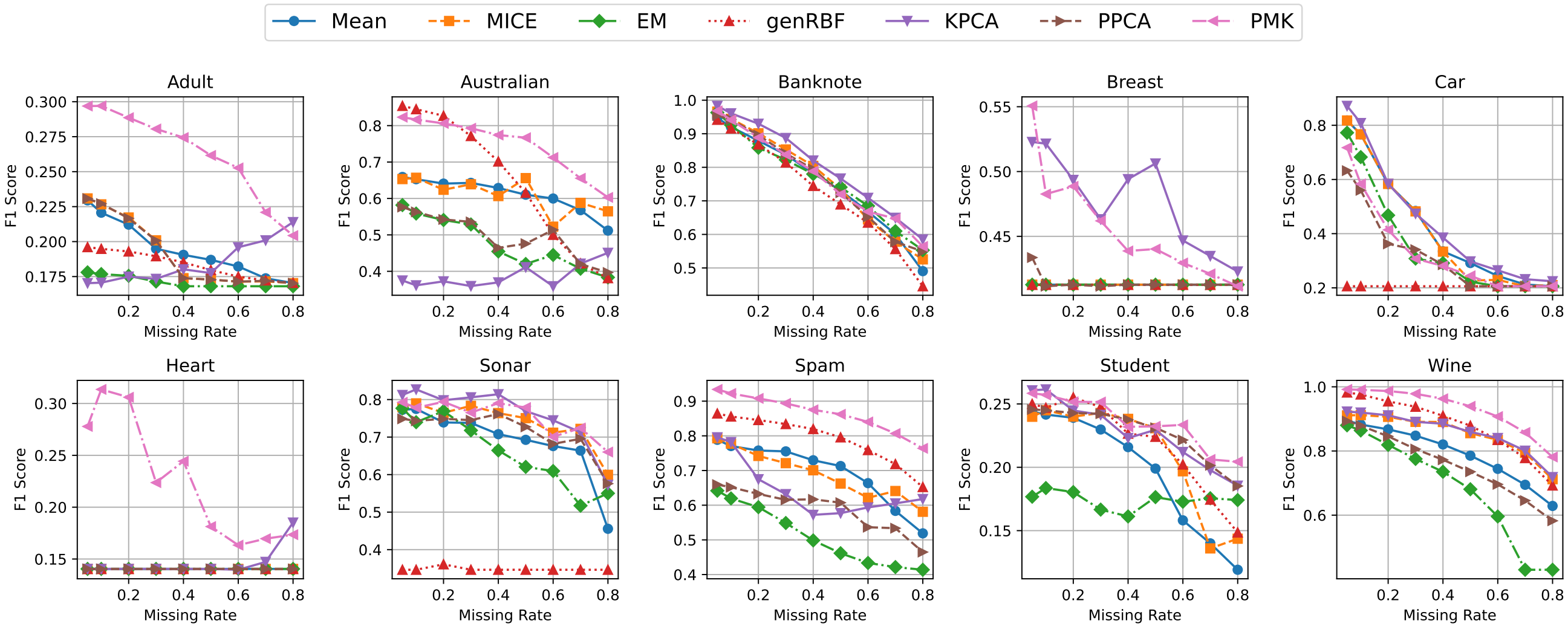

Figure 6: F1 scores for various models under the MCAR data mechanism across all 10 datasets

| Model | Hepatitis | Horse | Kidney | Mammo | Pima | Wisconsin |
|---|---|---|---|---|---|---|
| PMK | 0.1377 ± 0.0016 | 0.1066 ± ∼ 0.00 | **0.6164** ± ∼ 0.00 | **0.3271** ± ∼ 0.00 | **0.1374 ± 0.0019** | **0.7585 ± 0.0025** |
| Mean | 0.0015 ± ∼ 0.00 | 0.0078 ± ∼ 0.00 | 0.0067 ± ∼ 0.00 | 0.0959 ± ∼ 0.00 | 0.0111 ± ∼ 0.00 | 0.7295 ± ∼ 0.00 |
| MICE | 0.0021 ± 0.0012 | 0.0024 ± ∼ 0.00 | 0.0074 ± ∼ 0.00 | 0.0959 ± ∼ 0.00 | 0.0910 ± ∼ 0.00 | 0.7427 ± ∼ 0.00 |
| EM | 0.0025 ± 0.0019 | 0.0079 ± 0.0023 | 0.0045 ± 0.0017 | 0.0911 ± 0.0011 | 0.0211 ± 0.0050 | 0.7387 ± 0.0033 |
| RBF | 0.0772 ± 0.0052 | 0.0166 ± ∼ 0.00 | 0.0132 ± ∼ 0.00 | 0.0002 ± ∼ 0.00 | 0.0052 ± 0.0004 | 0.4505 ± ∼ 0.00 |
| KPCA | 0.0159 ± 0.0226 | 0.0583 ± 0.0175 | 0.0056 ± 0.0065 | 0.0117 ± 0.0111 | 0.0092 ± 0.0115 | 0.5186 ± 0.0672 |
| PPCA | 0.0015 ± ∼ 0.00 | 0.0540 ± 0.0047 | 0.0081 ± 0.0015 | 0.0959 ± ∼ 0.00 | 0.0909 ± ∼ 0.00 | 0.7427 ± ∼ 0.00 |
| Simp | 0.0019 ± 0.0012 | 0.0001 ± 0.0001 | 0.0424 ± 0.0050 | 0.0951 ± 0.0017 | 0.0556 ± 0.0021 | 0.0000 ± ∼ 0.00 |
| Gow | **0.1968** ± ∼ 0.00 | **0.1280** ± ∼ 0.00 | 0.3899 ± ∼ 0.00 | 0.2970 ± ∼ 0.00 | 0.1069 ± 0.0032 | 0.6939 ± ∼ 0.00 |

Table 6: Complete Table 4 presents the Clustering NMI performance results with corresponding standard deviations across different methods and incomplete datasets.This table provides both the NMI values and their standard deviations, while underlining highlights the second-best NMI value. ∼ 0.00 denotes cases where the standard deviation approaches zero.

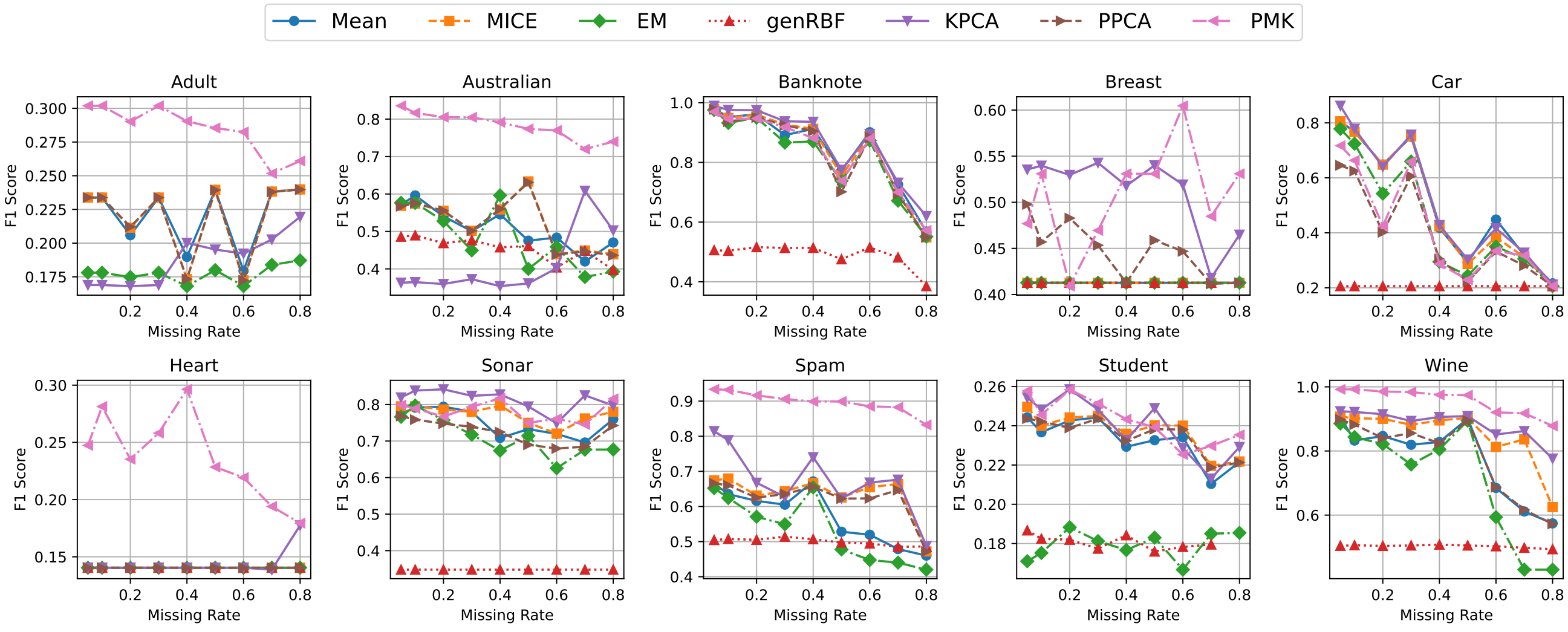

Figure 7: F1 scores for various models under the MAR data mechanism across all 10 datasets

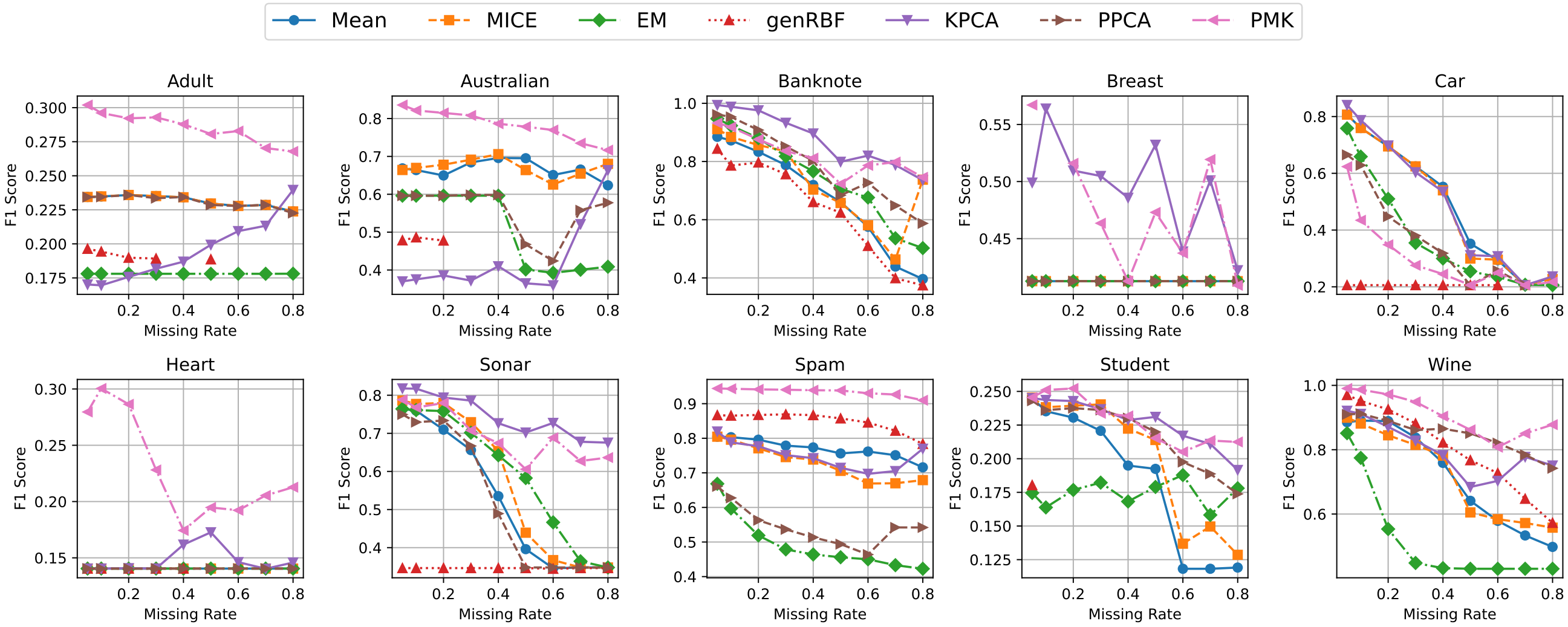

Figure 8: F1 scores for various models under the MNAR data mechanism across all 10 datasets

| Model | Hepatitis | Horse | Kidney | Mammo | Pima | Wisconsin |
|---|---|---|---|---|---|---|
| PMK | 0.1111 ± 0.0153 | 0.0049 ± 0.0186 | **0.6391 ± ∼ 0.00** | **0.4197 ± ∼ 0.00** | 0.1541 ± 0.0066 | **0.8463 ± 0.0026** |
| Mean | 0.0198 ± ∼ 0.00 | 0.0262 ± ∼ 0.00 | -0.0185 ± ∼ 0.00 | 0.1133 ± ∼ 0.00 | 0.0300 ± ∼ 0.00 | 0.8337 ± ∼ 0.00 |
| MICE | 0.0234 ± 0.0072 | 0.0110 ± ∼ 0.00 | -0.0184 ± 0.0013 | 0.1133 ± ∼ 0.00 | 0.1625 ± ∼ 0.00 | 0.8444 ± ∼ 0.00 |
| EM | 0.0248 ± 0.0099 | 0.0253 ± 0.0053 | -0.0043 ± 0.0017 | 0.1083 ± 0.0011 | 0.0572 ± 0.0086 | 0.8412 ± 0.0026 |
| RBF | 0.0834 ± 0.0034 | -0.0044 ± ∼ 0.00 | -0.0077 ± ∼ 0.00 | -0.0007 ± ∼ 0.00 | 0.0208 ± 0.0010 | 0.5390 ± ∼ 0.00 |
| KPCA | -0.0394 ± 0.0312 | 0.1060 ± 0.0284 | -0.0051 ± 0.0123 | 0.0138 ± 0.0135 | -0.0137 ± 0.0115 | 0.5001 ± 0.1082 |
| PPCA | 0.0198 ± ∼ 0.00 | 0.0250 ± 0.0052 | -0.0193 ± 0.0004 | 0.1133 ± ∼ 0.00 | 0.1621 ± 0.0004 | 0.8444 ± ∼ 0.00 |
| Sim | 0.0127 ± 0.0064 | -0.0030 ± 0.0010 | -0.0237 ± 0.0024 | 0.1101 ± 0.0066 | 0.0998 ± 0.0034 | 0.0000 ± ∼ 0.00 |
| Gow | **0.2115 ± ∼ 0.00** | **0.1341 ± ∼ 0.00** | 0.2880 ± ∼ 0.00 | 0.3548 ± ∼ 0.00 | **0.1753 ± 0.0038** | 0.8020 ± ∼ 0.00 |

Table 7: Clustering ARI performance results with corresponding standard deviations across different methods and incomplete datasets.This table provides both the ARI values and their standard deviations, while underlining highlights the second-best ARI value. ∼ 0.00 denotes cases where the standard deviation approaches zero.

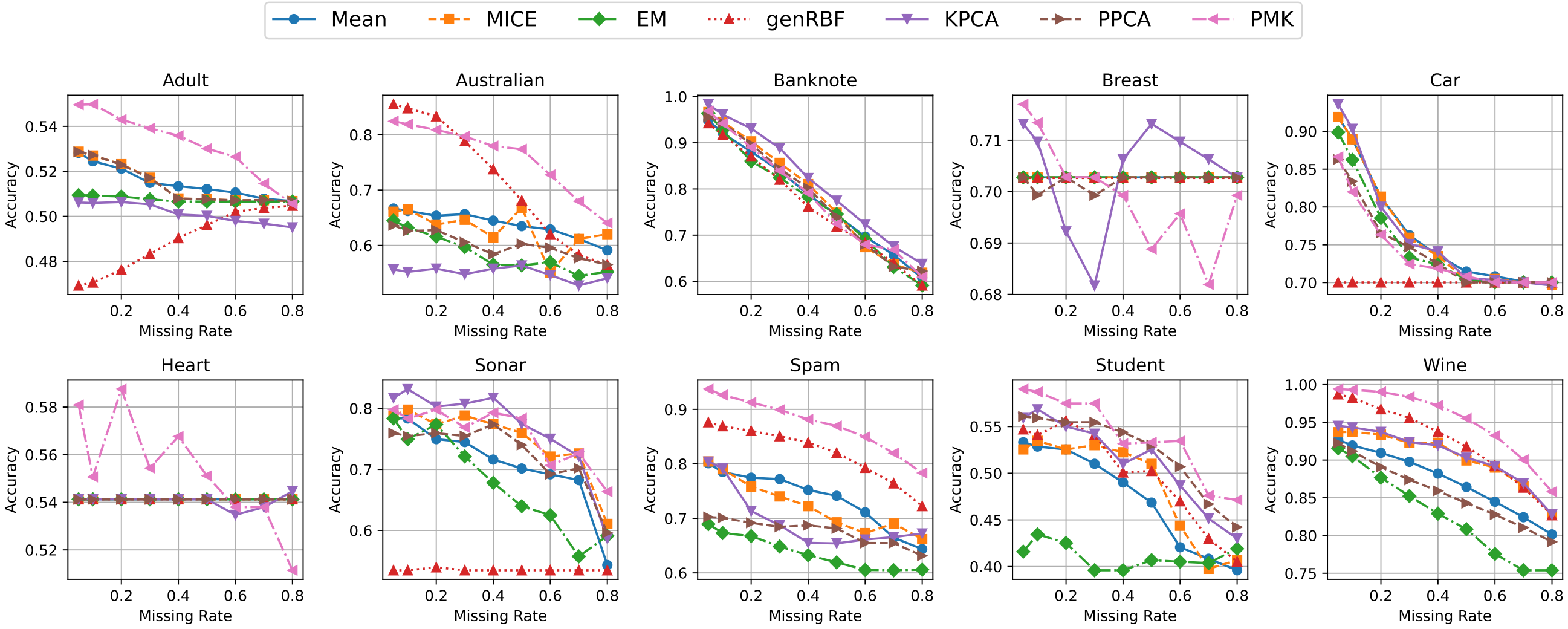

Figure 9: Accuracy scores for various models under the MCAR data mechanism across all 10 datasets

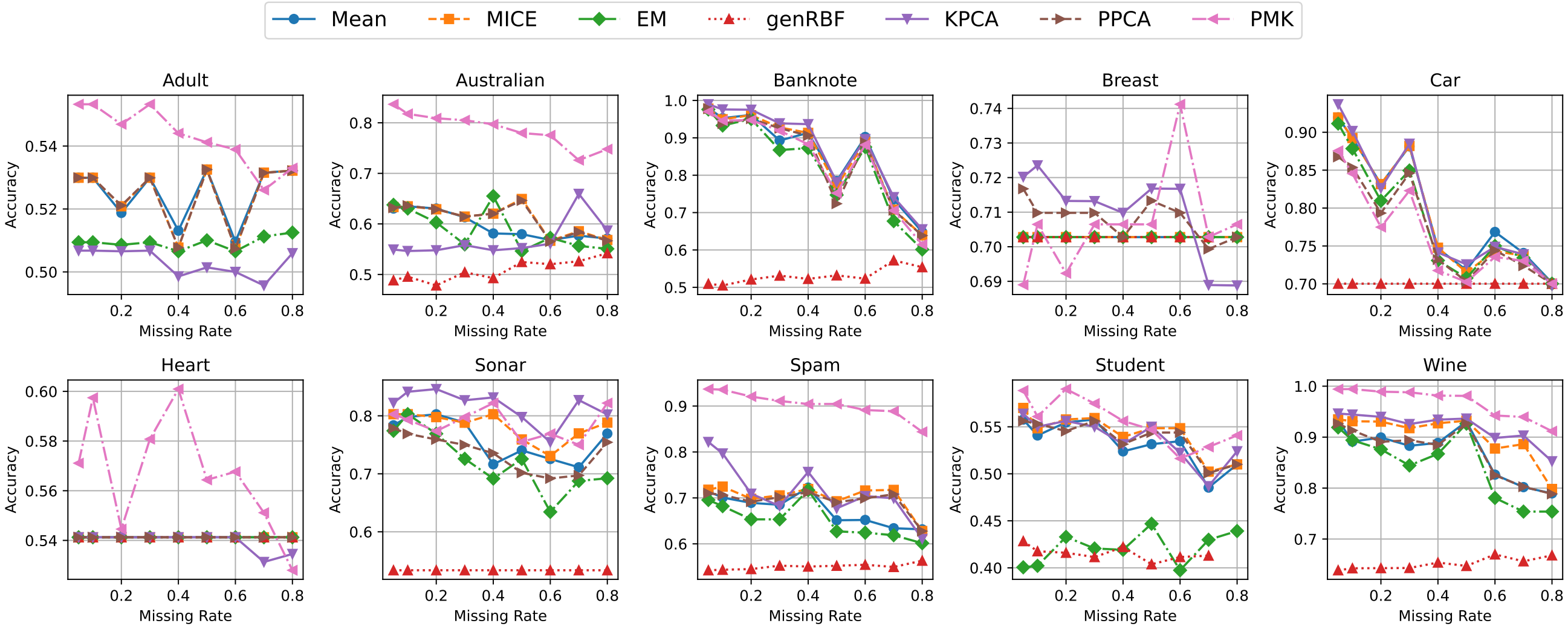

Figure 10: Accuracy scores for various models under the MAR data mechanism across all 10 datasets

| Model | Hepatitis | Horse | Kidney | Mammo | Pima | Wisconsin |
|---|---|---|---|---|---|---|
| PMK | 0.8065 ± 0.0353 | **0.8506 ± 0.0217** | **0.9875 ± 0.0137** | **0.8241 ± 0.0714** | 0.7618 ± 0.0312 | **0.9700 ± 0.0165** |
| Mean | 0.8129 ± 0.6991 | 0.8398 ± 0.8262 | 0.9775 ± 0.9761 | 0.8148 ± 0.8095 | **0.7735 ± 0.7324** | 0.9628 ± 0.9589 |
| MICE | 0.8129 ± 0.6938 | **0.8506 ± 0.8371** | 0.9700 ± 0.9682 | **0.8241 ± 0.8196** | 0.7722 ± 0.7317 | 0.9614 ± 0.9574 |
| EM | **0.8387 ± 0.7311** | 0.8424 ± 0.8300 | 0.9375 ± 0.9340 | 0.8210 ± 0.8165 | 0.7630 ± 0.7229 | 0.9643 ± 0.9605 |
| RBF | 0.7935 ± 0.4424 | 0.6304 ± 0.3867 | 0.6250 ± 0.3846 | 0.5140 ± 0.5123 | 0.6510 ± 0.3943 | 0.5564 ± 0.5024 |
| KPCA | 0.7935 ± 0.4424 | 0.6850 ± 0.5667 | 0.6250 ± 0.3846 | 0.8127 ± 0.8116 | 0.6510 ± 0.3943 | 0.9500 ± 0.9463 |
| PPCA | 0.8000 ± 0.6865 | **0.8506 ± 0.8363** | 0.9625 ± 0.9602 | **0.8241 ± 0.8196** | 0.7722 ± 0.7317 | 0.9628 ± 0.9589 |

Table 8: Complete Table 3 presents the Classification Accuracy results with corresponding standard deviations across different methods and incomplete datasets. Bold font is used to indicate the method with the highest accuracy for each dataset. In cases where there is a tie for the highest accuracy, a double underline is used to highlight the method with the smallest standard deviation.

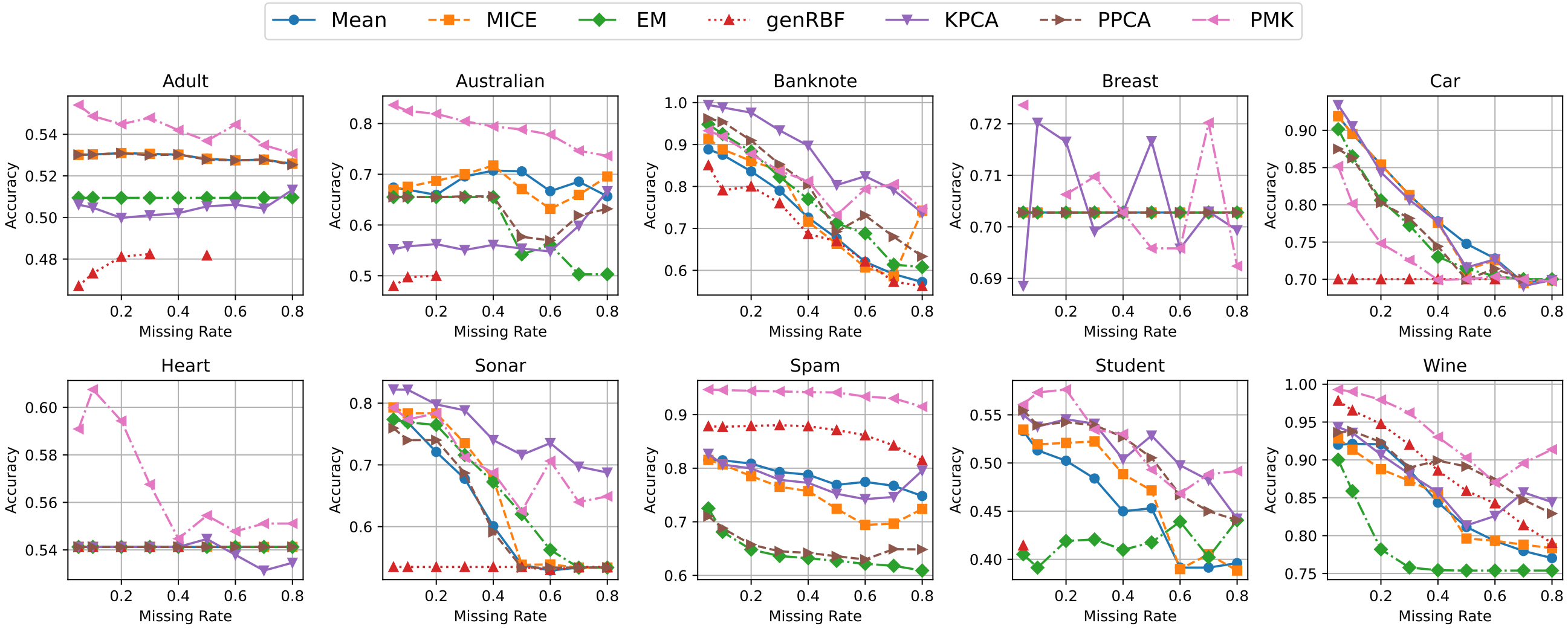

Figure 11: Accuracy scores for various models under the MNAR data mechanism across all 10 datasets

| Mechanism | Model | 0.05 | 0.1 | 0.2 | 0.3 | 0.4 | 0.5 | 0.6 | 0.7 | 0.8 |
|---|---|---|---|---|---|---|---|---|---|---|
| MCAR | Mean | $0.6590 \pm 0.040$ | $0.6528 \pm 0.027$ | $0.6408 \pm 0.035$ | $0.6428 \pm 0.043$ | $0.6287 \pm 0.026$ | $0.6103 \pm 0.036$ | $0.5999 \pm 0.043$ | $0.5679 \pm 0.059$ | $0.5118 \pm 0.046$ |
| | MICE | $0.6532 \pm 0.033$ | $0.6568 \pm 0.033$ | $0.6239 \pm 0.038$ | $0.6390 \pm 0.041$ | $0.6066 \pm 0.047$ | $0.6559 \pm 0.038$ | $0.5226 \pm 0.035$ | $0.5879 \pm 0.047$ | $0.5647 \pm 0.047$ |
| | EM | $0.5818 \pm 0.019$ | $0.5586 \pm 0.027$ | $0.5408 \pm 0.051$ | $0.5286 \pm 0.035$ | $0.4545 \pm 0.063$ | $0.4201 \pm 0.073$ | $0.4448 \pm 0.061$ | $0.4071 \pm 0.042$ | $0.3833 \pm 0.040$ |
| | genRBF | $0.8541 \pm 0.030$ | $0.8451 \pm 0.030$ | $0.8269 \pm 0.018$ | $0.7714 \pm 0.022$ | $0.7014 \pm 0.029$ | $0.6155 \pm 0.032$ | $0.5002 \pm 0.036$ | $0.4216 \pm 0.024$ | $0.3812 \pm 0.019$ |
| | KPCA | $0.3748 \pm 0.008$ | $0.3615 \pm 0.014$ | $0.3724 \pm 0.019$ | $0.3595 \pm 0.006$ | $0.3696 \pm 0.017$ | $0.4113 \pm 0.034$ | $0.3590 \pm 0.007$ | $0.4210 \pm 0.046$ | $0.4508 \pm 0.042$ |
| | PPCA | $0.5764 \pm 0.021$ | $0.5631 \pm 0.023$ | $0.5424 \pm 0.068$ | $0.5356 \pm 0.043$ | $0.4645 \pm 0.039$ | $0.4759 \pm 0.103$ | $0.5138 \pm 0.054$ | $0.4198 \pm 0.036$ | $0.3968 \pm 0.035$ |
| | PMK | $0.8231 \pm 0.031$ | $0.8166 \pm 0.033$ | $0.8061 \pm 0.039$ | $0.7928 \pm 0.043$ | $0.7741 \pm 0.039$ | $0.7668 \pm 0.040$ | $0.7125 \pm 0.028$ | $0.6554 \pm 0.045$ | $0.6029 \pm 0.025$ |
| MAR | Mean | $0.5711 \pm 0.033$ | $0.5961 \pm 0.030$ | $0.5404 \pm 0.035$ | $0.5008 \pm 0.022$ | $0.5457 \pm 0.082$ | $0.4754 \pm 0.055$ | $0.4836 \pm 0.054$ | $0.4198 \pm 0.029$ | $0.4706 \pm 0.037$ |
| | MICE | $0.5685 \pm 0.026$ | $0.5744 \pm 0.039$ | $0.5553 \pm 0.027$ | $0.5018 \pm 0.022$ | $0.5588 \pm 0.057$ | $0.6335 \pm 0.037$ | $0.4395 \pm 0.051$ | $0.4492 \pm 0.065$ | $0.4392 \pm 0.035$ |
| | EM | $0.5765 \pm 0.021$ | $0.5761 \pm 0.025$ | $0.5278 \pm 0.037$ | $0.4494 \pm 0.053$ | $0.5962 \pm 0.017$ | $0.4000 \pm 0.059$ | $0.4588 \pm 0.079$ | $0.3783 \pm 0.025$ | $0.3928 \pm 0.037$ |
| | genRBF | $0.4859 \pm 0.042$ | $0.4895 \pm 0.040$ | $0.4688 \pm 0.028$ | $0.4767 \pm 0.036$ | $0.4576 \pm 0.056$ | $0.4611 \pm 0.044$ | $0.4039 \pm 0.023$ | $0.4483 \pm 0.033$ | $0.3970 \pm 0.029$ |
| | KPCA | $0.3631 \pm 0.009$ | $0.3646 \pm 0.006$ | $0.3597 \pm 0.010$ | $0.3726 \pm 0.011$ | $0.3539 \pm 0.003$ | $0.3612 \pm 0.011$ | $0.4018 \pm 0.047$ | $0.6086 \pm 0.070$ | $0.5028 \pm 0.103$ |
| | PPCA | $0.5674 \pm 0.026$ | $0.5755 \pm 0.038$ | $0.5553 \pm 0.027$ | $0.5018 \pm 0.022$ | $0.5598 \pm 0.057$ | $0.6310 \pm 0.039$ | $0.4382 \pm 0.048$ | $0.4471 \pm 0.061$ | $0.4361 \pm 0.037$ |
| | PMK | $0.8355 \pm 0.019$ | $0.8164 \pm 0.024$ | $0.8252 \pm 0.026$ | $0.8048 \pm 0.010$ | $0.7915 \pm 0.026$ | $0.7738 \pm 0.024$ | $0.7697 \pm 0.048$ | $0.7198 \pm 0.019$ | $0.7399 \pm 0.041$ |
| MNAR | Mean | $0.6684 \pm 0.021$ | $0.6639 \pm 0.018$ | $0.6496 \pm 0.026$ | $0.6842 \pm 0.015$ | $0.6959 \pm 0.033$ | $0.6952 \pm 0.036$ | $0.6509 \pm 0.049$ | $0.6653 \pm 0.036$ | $0.6236 \pm 0.038$ |
| | MICE | $0.6642 \pm 0.023$ | $0.6697 \pm 0.020$ | $0.6780 \pm 0.041$ | $0.6914 \pm 0.012$ | $0.7060 \pm 0.032$ | $0.6639 \pm 0.014$ | $0.6257 \pm 0.015$ | $0.6545 \pm 0.048$ | $0.6803 \pm 0.024$ |
| | EM | $0.5962 \pm 0.017$ | $0.5962 \pm 0.017$ | $0.5962 \pm 0.017$ | $0.5962 \pm 0.017$ | $0.5962 \pm 0.017$ | $0.4014 \pm 0.051$ | $0.3925 \pm 0.030$ | $0.4006 \pm 0.040$ | $0.4086 \pm 0.059$ |
| | genRBF | $0.4784 \pm 0.043$ | $0.4862 \pm 0.039$ | $0.4778 \pm 0.032$ | - | - | - | - | - | - |
| | KPCA | $0.3699 \pm 0.010$ | $0.3753 \pm 0.012$ | $0.3855 \pm 0.013$ | $0.3720 \pm 0.013$ | $0.4098 \pm 0.018$ | $0.3649 \pm 0.013$ | $0.3595 \pm 0.005$ | $0.5211 \pm 0.133$ | $0.6637 \pm 0.045$ |
| | PPCA | $0.5962 \pm 0.017$ | $0.5962 \pm 0.017$ | $0.5962 \pm 0.017$ | $0.5983 \pm 0.020$ | $0.5983 \pm 0.020$ | $0.4676 \pm 0.018$ | $0.4240 \pm 0.035$ | $0.5578 \pm 0.087$ | $0.5777 \pm 0.012$ |
| | PMK | $0.8357 \pm 0.012$ | $0.8214 \pm 0.016$ | $0.8149 \pm 0.033$ | $0.8082 \pm 0.039$ | $0.7863 \pm 0.056$ | $0.7785 \pm 0.046$ | $0.7696 \pm 0.040$ | $0.7349 \pm 0.027$ | $0.7166 \pm 0.029$ |

Table 9: F1 Scores with corresponding standard deviations across various models and missing value proportions for the Australian Dataset.

| Mechanism | Model | 0.05 | .1 | .2 | .3 | .4 | .5 | .6 | .7 | .8 |
|---|---|---|---|---|---|---|---|---|---|---|
| MCAR | Mean | 0.9475 ± 0.006 | 0.9201 ± 0.009 | 0.8791 ± 0.007 | 0.8376 ± 0.012 | 0.7824 ± 0.011 | 0.7309 ± 0.015 | 0.6679 ± 0.037 | 0.6007 ± 0.047 | 0.4905 ± 0.041 |
| | MICE | 0.9660 ± 0.007 | 0.9444 ± 0.010 | 0.9014 ± 0.005 | 0.8532 ± 0.005 | 0.8041 ± 0.007 | 0.7350 ± 0.010 | 0.6445 ± 0.034 | 0.5782 ± 0.037 | 0.5256 ± 0.032 |
| | EM | 0.9631 ± 0.003 | 0.9262 ± 0.016 | 0.8582 ± 0.031 | 0.8223 ± 0.020 | 0.7798 ± 0.013 | 0.7407 ± 0.027 | 0.6848 ± 0.024 | 0.6099 ± 0.019 | 0.5529 ± 0.041 |
| | genRBF | 0.9416 ± 0.004 | 0.9156 ± 0.008 | 0.8679 ± 0.010 | 0.8143 ± 0.015 | 0.7445 ± 0.034 | 0.6893 ± 0.030 | 0.6350 ± 0.047 | 0.5563 ± 0.042 | 0.4455 ± 0.062 |
| | KPCA | 0.9830 ± 0.008 | 0.9609 ± 0.006 | 0.9298 ± 0.020 | 0.8873 ± 0.019 | 0.8203 ± 0.037 | 0.7666 ± 0.031 | 0.7090 ± 0.013 | 0.6495 ± 0.017 | 0.5851 ± 0.039 |
| | PPCA | 0.9547 ± 0.020 | 0.9422 ± 0.012 | 0.8957 ± 0.017 | 0.8410 ± 0.014 | 0.7944 ± 0.022 | 0.7243 ± 0.020 | 0.6524 ± 0.015 | 0.5777 ± 0.034 | 0.5494 ± 0.040 |
| | MPK | 0.9683 ± 0.011 | 0.9411 ± 0.021 | 0.8885 ± 0.023 | 0.8377 ± 0.010 | 0.7883 ± 0.023 | 0.7194 ± 0.021 | 0.6680 ± 0.015 | 0.6472 ± 0.015 | 0.5641 ± 0.038 |
| MAR | Mean | 0.9794 ± 0.006 | 0.9521 ± 0.010 | 0.9610 ± 0.007 | 0.8914 ± 0.015 | 0.9124 ± 0.008 | 0.7725 ± 0.029 | 0.9012 ± 0.014 | 0.7258 ± 0.033 | 0.5683 ± 0.014 |
| | MICE | 0.9779 ± 0.006 | 0.9490 ± 0.014 | 0.9601 ± 0.010 | 0.9265 ± 0.014 | 0.9126 ± 0.011 | 0.7562 ± 0.026 | 0.8881 ± 0.011 | 0.6953 ± 0.037 | 0.5488 ± 0.027 |
| | EM | 0.9757 ± 0.007 | 0.9322 ± 0.011 | 0.9499 ± 0.019 | 0.8664 ± 0.018 | 0.8701 ± 0.017 | 0.7382 ± 0.038 | 0.8737 ± 0.024 | 0.6717 ± 0.033 | 0.5514 ± 0.036 |
| | genRBF | 0.5058 ± 0.018 | 0.5038 ± 0.014 | 0.5161 ± 0.033 | 0.5132 ± 0.021 | 0.5137 ± 0.019 | 0.4755 ± 0.022 | 0.5150 ± 0.019 | 0.4815 ± 0.015 | 0.3853 ± 0.021 |
| | KPCA | 0.9897 ± 0.004 | 0.9756 ± 0.008 | 0.9749 ± 0.016 | 0.9377 ± 0.007 | 0.9360 ± 0.014 | 0.7751 ± 0.025 | 0.8952 ± 0.013 | 0.7320 ± 0.028 | 0.6201 ± 0.024 |
| | PPCA | 0.9779 ± 0.006 | 0.9369 ± 0.023 | 0.9504 ± 0.014 | 0.9243 ± 0.016 | 0.9052 ± 0.015 | 0.7015 ± 0.027 | 0.8850 ± 0.009 | 0.6941 ± 0.037 | 0.5491 ± 0.028 |
| | MPK | 0.9698 ± 0.008 | 0.9454 ± 0.016 | 0.9470 ± 0.009 | 0.9182 ± 0.010 | 0.8813 ± 0.014 | 0.7381 ± 0.029 | 0.8795 ± 0.011 | 0.6997 ± 0.033 | 0.5722 ± 0.043 |
| MNAR | Mean | 0.8848 ± 0.018 | 0.8724 ± 0.016 | 0.8335 ± 0.012 | 0.7877 ± 0.014 | 0.7193 ± 0.011 | 0.6606 ± 0.021 | 0.5758 ± 0.031 | 0.4390 ± 0.024 | 0.3962 ± 0.020 |
| | MICE | 0.9112 ± 0.013 | 0.8846 ± 0.017 | 0.8564 ± 0.022 | 0.8353 ± 0.025 | 0.7036 ± 0.022 | 0.6575 ± 0.015 | 0.5809 ± 0.028 | 0.4635 ± 0.026 | 0.7376 ± 0.030 |
| | EM | 0.9471 ± 0.016 | 0.9233 ± 0.013 | 0.8804 ± 0.017 | 0.8179 ± 0.026 | 0.7663 ± 0.016 | 0.7059 ± 0.005 | 0.6762 ± 0.035 | 0.5362 ± 0.040 | 0.5025 ± 0.025 |
| | genRBF | 0.8439 ± 0.019 | 0.7860 ± 0.018 | 0.7967 ± 0.018 | 0.7562 ± 0.035 | 0.6606 ± 0.045 | 0.6242 ± 0.039 | 0.5099 ± 0.065 | 0.3990 ± 0.025 | 0.3745 ± 0.018 |
| | KPCA | 0.9941 ± 0.006 | 0.9882 ± 0.006 | 0.9757 ± 0.009 | 0.9331 ± 0.013 | 0.8963 ± 0.019 | 0.7990 ± 0.034 | 0.8202 ± 0.029 | 0.7879 ± 0.017 | 0.7357 ± 0.023 |
| | PPCA | 0.9614 ± 0.016 | 0.9533 ± 0.014 | 0.9081 ± 0.027 | 0.8516 ± 0.014 | 0.8025 ± 0.019 | 0.6838 ± 0.023 | 0.7268 ± 0.034 | 0.6486 ± 0.115 | 0.5873 ± 0.101 |
| | MPK | 0.9322 ± 0.019 | 0.9183 ± 0.015 | 0.8760 ± 0.015 | 0.8357 ± 0.019 | 0.8118 ± 0.017 | 0.7233 ± 0.025 | 0.7865 ± 0.039 | 0.7981 ± 0.048 | 0.7458 ± 0.035 |

Table 10: F1 Scores with corresponding standard deviations across various models and missing value proportions for the Banknote dataset.

| Mechanism | Model | 0.05 | 0.1 | 0.2 | 0.3 | 0.4 | 0.5 | 0.6 | 0.7 | 0.8 |
|---|---|---|---|---|---|---|---|---|---|---|
| MCAR | Mean | 0.4127 ± 0.001 | 0.4127 ± 0.001 | 0.4127 ± 0.001 | 0.4127 ± 0.001 | 0.4127 ± 0.001 | 0.4127 ± 0.001 | 0.4127 ± 0.001 | 0.4127 ± 0.001 | 0.4127 ± 0.001 |
| | MICE | 0.4127 ± 0.001 | 0.4127 ± 0.001 | 0.4127 ± 0.001 | 0.4127 ± 0.001 | 0.4127 ± 0.001 | 0.4127 ± 0.001 | 0.4127 ± 0.001 | 0.4127 ± 0.001 | 0.4127 ± 0.001 |
| | EM | 0.4127 ± 0.001 | 0.4127 ± 0.001 | 0.4127 ± 0.001 | 0.4127 ± 0.001 | 0.4127 ± 0.001 | 0.4127 ± 0.001 | 0.4127 ± 0.001 | 0.4127 ± 0.001 | 0.4127 ± 0.001 |
| | genRBF | 0.4126 ± 0.008 | 0.4126 ± 0.008 | 0.4126 ± 0.008 | 0.4126 ± 0.008 | 0.4126 ± 0.008 | 0.4126 ± 0.008 | 0.4126 ± 0.008 | 0.4126 ± 0.008 | 0.4126 ± 0.008 |
| | KPCA | 0.5228 ± 0.076 | 0.5215 ± 0.074 | 0.4935 ± 0.065 | 0.4632 ± 0.064 | 0.4941 ± 0.051 | 0.5062 ± 0.050 | 0.4468 ± 0.028 | 0.4349 ± 0.027 | 0.4229 ± 0.020 |
| | PPCA | 0.4336 ± 0.029 | 0.4115 ± 0.002 | 0.4127 ± 0.001 | 0.4115 ± 0.003 | 0.4127 ± 0.001 | 0.4127 ± 0.001 | 0.4127 ± 0.001 | 0.4127 ± 0.001 | 0.4127 ± 0.001 |
| | MPK | 0.5506 ± 0.088 | 0.4826 ± 0.081 | 0.4886 ± 0.079 | 0.4619 ± 0.066 | 0.4387 ± 0.032 | 0.4402 ± 0.044 | 0.4295 ± 0.044 | 0.4211 ± 0.028 | 0.4115 ± 0.003 |
| MAR | Mean | 0.4127 ± 0.001 | 0.4127 ± 0.001 | 0.4127 ± 0.001 | 0.4127 ± 0.001 | 0.4127 ± 0.001 | 0.4127 ± 0.001 | 0.4127 ± 0.001 | 0.4127 ± 0.001 | 0.4127 ± 0.001 |
| | MICE | 0.4127 ± 0.001 | 0.4127 ± 0.001 | 0.4127 ± 0.001 | 0.4127 ± 0.001 | 0.4127 ± 0.001 | 0.4127 ± 0.001 | 0.4127 ± 0.001 | 0.4127 ± 0.001 | 0.4127 ± 0.001 |
| | EM | 0.4127 ± 0.001 | 0.4127 ± 0.001 | 0.4127 ± 0.001 | 0.4127 ± 0.001 | 0.4127 ± 0.001 | 0.4127 ± 0.001 | 0.4127 ± 0.001 | 0.4127 ± 0.001 | 0.4127 ± 0.001 |
| | genRBF | 0.4127 ± 0.001 | 0.4127 ± 0.001 | 0.4127 ± 0.001 | 0.4127 ± 0.001 | 0.4127 ± 0.001 | 0.4127 ± 0.001 | 0.4127 ± 0.001 | 0.4127 ± 0.001 | - |
| | KPCA | 0.5355 ± 0.087 | 0.5397 ± 0.098 | 0.5297 ± 0.045 | 0.5428 ± 0.033 | 0.5179 ± 0.054 | 0.5401 ± 0.034 | 0.5192 ± 0.050 | 0.4180 ± 0.023 | 0.4648 ± 0.059 |
| | PPCA | 0.4978 ± 0.075 | 0.4570 ± 0.023 | 0.4828 ± 0.063 | 0.4534 ± 0.059 | 0.4127 ± 0.001 | 0.4588 ± 0.023 | 0.4468 ± 0.028 | 0.4115 ± 0.002 | 0.4127 ± 0.001 |
| | MPK | 0.4769 ± 0.071 | 0.5309 ± 0.072 | 0.4091 ± 0.005 | 0.4694 ± 0.060 | 0.5309 ± 0.072 | 0.5309 ± 0.072 | 0.6046 ± 0.048 | 0.4848 ± 0.046 | 0.5309 ± 0.072 |
| MNAR | Mean | 0.4127 ± 0.001 | 0.4127 ± 0.001 | 0.4127 ± 0.001 | 0.4127 ± 0.001 | 0.4127 ± 0.001 | 0.4127 ± 0.001 | 0.4127 ± 0.001 | 0.4127 ± 0.001 | 0.4127 ± 0.001 |
| | MICE | 0.4127 ± 0.001 | 0.4127 ± 0.001 | 0.4127 ± 0.001 | 0.4127 ± 0.001 | 0.4127 ± 0.001 | 0.4127 ± 0.001 | 0.4127 ± 0.001 | 0.4127 ± 0.001 | 0.4127 ± 0.001 |
| | EM | 0.4127 ± 0.001 | 0.4127 ± 0.001 | 0.4127 ± 0.001 | 0.4127 ± 0.001 | 0.4127 ± 0.001 | 0.4127 ± 0.001 | 0.4127 ± 0.001 | 0.4127 ± 0.001 | 0.4127 ± 0.001 |
| | genRBF | - | - | - | - | - | - | - | - | - |
| | KPCA | 0.4991 ± 0.098 | 0.5639 ± 0.073 | 0.5096 ± 0.097 | 0.5048 ± 0.076 | 0.4858 ± 0.049 | 0.5322 ± 0.075 | 0.4387 ± 0.039 | 0.5010 ± 0.045 | 0.4223 ± 0.025 |
| | PPCA | 0.4127 ± 0.001 | 0.4127 ± 0.001 | 0.4127 ± 0.001 | 0.4127 ± 0.001 | 0.4127 ± 0.001 | 0.4127 ± 0.001 | 0.4127 ± 0.001 | 0.4127 ± 0.001 | 0.4127 ± 0.001 |
| | MPK | 0.5670 ± 0.027 | - | 0.5160 ± 0.085 | 0.4635 ± 0.044 | 0.4127 ± 0.001 | 0.4730 ± 0.075 | 0.4373 ± 0.033 | 0.5193 ± 0.052 | 0.4091 ± 0.004 |

Table 11: F1 Scores with corresponding standard deviations across various models and missing value proportions for the Breast dataset.

| Mechanism | Model | 0.05 | 0.1 | 0.2 | 0.3 | 0.4 | 0.5 | 0.6 | 0.7 | 0.8 |
|---|---|---|---|---|---|---|---|---|---|---|
| MCAR | Mean | 0.7727 ± 0.060 | 0.7751 ± 0.047 | 0.7390 ± 0.065 | 0.7378 ± 0.061 | 0.7077 ± 0.028 | 0.6927 ± 0.067 | 0.6760 ± 0.056 | 0.6639 ± 0.061 | 0.4556 ± 0.024 |
| | MICE | 0.7798 ± 0.063 | 0.7898 ± 0.042 | 0.7647 ± 0.050 | 0.7833 ± 0.063 | 0.7639 ± 0.069 | 0.7504 ± 0.062 | 0.7117 ± 0.031 | 0.7219 ± 0.069 | 0.5994 ± 0.041 |
| | EM | 0.7768 ± 0.038 | 0.7396 ± 0.076 | 0.7692 ± 0.073 | 0.7179 ± 0.023 | 0.6645 ± 0.025 | 0.6205 ± 0.040 | 0.6095 ± 0.050 | 0.5172 ± 0.052 | 0.5501 ± 0.017 |
| | genRBF | 0.3464 ± 0.036 | 0.3464 ± 0.036 | 0.3609 ± 0.060 | 0.3464 ± 0.036 | 0.3464 ± 0.036 | 0.3464 ± 0.036 | 0.3464 ± 0.036 | 0.3464 ± 0.036 | 0.3464 ± 0.036 |
| | KPCA | 0.8122 ± 0.048 | 0.8271 ± 0.048 | 0.7981 ± 0.047 | 0.8059 ± 0.064 | 0.8141 ± 0.053 | 0.7692 ± 0.071 | 0.7446 ± 0.042 | 0.7128 ± 0.070 | 0.5727 ± 0.041 |
| | PPCA | 0.7486 ± 0.056 | 0.7432 ± 0.047 | 0.7487 ± 0.069 | 0.7452 ± 0.055 | 0.7615 ± 0.079 | 0.7272 ± 0.056 | 0.6819 ± 0.044 | 0.6955 ± 0.047 | 0.5758 ± 0.068 |
| | MPK | 0.7927 ± 0.045 | 0.7781 ± 0.049 | 0.7940 ± 0.056 | 0.7653 ± 0.044 | 0.7903 ± 0.049 | 0.7787 ± 0.090 | 0.7007 ± 0.055 | 0.7242 ± 0.036 | 0.6602 ± 0.046 |
| MAR | Mean | 0.7744 ± 0.054 | 0.7915 ± 0.037 | 0.7941 ± 0.068 | 0.7799 ± 0.055 | 0.7084 ± 0.029 | 0.7326 ± 0.058 | 0.7196 ± 0.068 | 0.6960 ± 0.061 | 0.7598 ± 0.071 |
| | MICE | 0.7950 ± 0.062 | 0.7950 ± 0.043 | 0.7864 ± 0.071 | 0.7798 ± 0.077 | 0.7971 ± 0.042 | 0.7498 ± 0.054 | 0.7204 ± 0.035 | 0.7619 ± 0.070 | 0.7801 ± 0.081 |
| | EM | 0.7663 ± 0.053 | 0.7973 ± 0.045 | 0.7573 ± 0.065 | 0.7175 ± 0.046 | 0.6737 ± 0.067 | 0.7144 ± 0.041 | 0.6257 ± 0.051 | 0.6766 ± 0.042 | 0.6767 ± 0.055 |
| | genRBF | 0.3480 ± 0.004 | 0.3480 ± 0.004 | 0.3480 ± 0.004 | 0.3480 ± 0.004 | 0.3480 ± 0.004 | 0.3480 ± 0.004 | 0.3480 ± 0.004 | 0.3480 ± 0.004 | 0.3480 ± 0.004 |
| | KPCA | 0.8192 ± 0.069 | 0.8382 ± 0.042 | 0.8416 ± 0.078 | 0.8237 ± 0.068 | 0.8278 ± 0.057 | 0.7947 ± 0.052 | 0.7484 ± 0.057 | 0.8251 ± 0.071 | 0.7971 ± 0.098 |
| | PPCA | 0.7685 ± 0.081 | 0.7587 ± 0.048 | 0.7476 ± 0.052 | 0.7389 ± 0.097 | 0.7245 ± 0.069 | 0.6898 ± 0.042 | 0.6796 ± 0.066 | 0.6845 ± 0.100 | 0.7427 ± 0.087 |
| | MPK | 0.7985 ± 0.038 | 0.7892 ± 0.034 | 0.7669 ± 0.074 | 0.7936 ± 0.061 | 0.8173 ± 0.035 | 0.7500 ± 0.088 | 0.7608 ± 0.032 | 0.7452 ± 0.090 | 0.8157 ± 0.049 |
| MNAR | Mean | 0.7641 ± 0.071 | 0.7595 ± 0.060 | 0.7096 ± 0.057 | 0.6560 ± 0.059 | 0.5352 ± 0.080 | 0.3960 ± 0.044 | 0.3459 ± 0.005 | 0.3480 ± 0.004 | 0.3480 ± 0.004 |
| | MICE | 0.7859 ± 0.052 | 0.7771 ± 0.045 | 0.7796 ± 0.062 | 0.7292 ± 0.049 | 0.6535 ± 0.077 | 0.4392 ± 0.069 | 0.3673 ± 0.044 | 0.3480 ± 0.004 | 0.3480 ± 0.004 |
| | EM | 0.7641 ± 0.049 | 0.7610 ± 0.077 | 0.7583 ± 0.065 | 0.7017 ± 0.102 | 0.6424 ± 0.087 | 0.5826 ± 0.094 | 0.4664 ± 0.065 | 0.3642 ± 0.041 | 0.3480 ± 0.004 |
| | genRBF | 0.3464 ± 0.036 | 0.3464 ± 0.036 | 0.3464 ± 0.036 | 0.3464 ± 0.036 | 0.3464 ± 0.036 | 0.3464 ± 0.036 | 0.3446 ± 0.034 | 0.3464 ± 0.036 | 0.3464 ± 0.036 |
| | KPCA | 0.8178 ± 0.041 | 0.8172 ± 0.055 | 0.7939 ± 0.041 | 0.7859 ± 0.078 | 0.7265 ± 0.061 | 0.7020 ± 0.070 | 0.7273 ± 0.031 | 0.6779 ± 0.051 | 0.6755 ± 0.038 |
| | PPCA | 0.7495 ± 0.049 | 0.7294 ± 0.058 | 0.7335 ± 0.057 | 0.6669 ± 0.076 | 0.4894 ± 0.096 | 0.3480 ± 0.004 | 0.3480 ± 0.004 | 0.3480 ± 0.004 | 0.3480 ± 0.004 |
| | MPK | 0.7883 ± 0.051 | 0.7677 ± 0.053 | 0.7797 ± 0.050 | 0.7075 ± 0.061 | 0.6742 ± 0.047 | 0.6058 ± 0.062 | 0.6885 ± 0.055 | 0.6274 ± 0.060 | 0.6364 ± 0.049 |

Table 12: F1 Scores with corresponding standard deviations across various models and missing value proportions for the Sonar dataset.

| Mechanism | Model | 0.05 | 0.1 | 0.2 | 0.3 | 0.4 | 0.5 | 0.6 | 0.7 | 0.8 |
|---|---|---|---|---|---|---|---|---|---|---|
| MCAR | Mean | 0.7886 ± 0.012 | 0.7706 ± 0.011 | 0.7576 ± 0.010 | 0.7552 ± 0.018 | 0.7296 ± 0.019 | 0.7129 ± 0.019 | 0.6636 ± 0.029 | 0.5836 ± 0.023 | 0.5188 ± 0.025 |
| | MICE | 0.7930 ± 0.013 | 0.7775 ± 0.013 | 0.7420 ± 0.010 | 0.7213 ± 0.015 | 0.7007 ± 0.023 | 0.6623 ± 0.018 | 0.6213 ± 0.020 | 0.6409 ± 0.019 | 0.5810 ± 0.021 |
| | EM | 0.6418 ± 0.013 | 0.6195 ± 0.015 | 0.5943 ± 0.026 | 0.5487 ± 0.025 | 0.4985 ± 0.038 | 0.4615 ± 0.038 | 0.4335 ± 0.031 | 0.4216 ± 0.021 | 0.4138 ± 0.030 |
| | genRBF | 0.8644 ± 0.021 | 0.8557 ± 0.023 | 0.8456 ± 0.022 | 0.8340 ± 0.025 | 0.8196 ± 0.030 | 0.7956 ± 0.028 | 0.7589 ± 0.018 | 0.7185 ± 0.017 | 0.6524 ± 0.013 |
| | KPCA | 0.7958 ± 0.017 | 0.7821 ± 0.025 | 0.6749 ± 0.047 | 0.6319 ± 0.053 | 0.5722 ± 0.024 | 0.5771 ± 0.056 | 0.5936 ± 0.057 | 0.6052 ± 0.074 | 0.6177 ± 0.067 |
| | PPCA | 0.6594 ± 0.018 | 0.6518 ± 0.017 | 0.6324 ± 0.018 | 0.6155 ± 0.030 | 0.6169 ± 0.031 | 0.6073 ± 0.032 | 0.5365 ± 0.058 | 0.5337 ± 0.022 | 0.4648 ± 0.024 |
| | MPK | 0.9339 ± 0.009 | 0.9222 ± 0.008 | 0.9080 ± 0.011 | 0.8935 ± 0.013 | 0.8750 ± 0.010 | 0.8622 ± 0.009 | 0.8401 ± 0.012 | 0.8067 ± 0.012 | 0.7639 ± 0.015 |
| MAR | Mean | 0.6648 ± 0.017 | 0.6352 ± 0.024 | 0.6160 ± 0.021 | 0.6052 ± 0.020 | 0.6715 ± 0.023 | 0.5282 ± 0.024 | 0.5195 ± 0.014 | 0.4788 ± 0.024 | 0.4602 ± 0.012 |
| | MICE | 0.6744 ± 0.017 | 0.6796 ± 0.023 | 0.6312 ± 0.031 | 0.6434 ± 0.023 | 0.6670 ± 0.025 | 0.6256 ± 0.017 | 0.6549 ± 0.010 | 0.6641 ± 0.036 | 0.4722 ± 0.037 |
| | EM | 0.6523 ± 0.019 | 0.6241 ± 0.017 | 0.5708 ± 0.033 | 0.5494 ± 0.024 | 0.6542 ± 0.016 | 0.4777 ± 0.046 | 0.4477 ± 0.025 | 0.4401 ± 0.029 | 0.4201 ± 0.058 |
| | genRBF | 0.5046 ± 0.012 | 0.5074 ± 0.009 | 0.5052 ± 0.011 | 0.5140 ± 0.007 | 0.5068 ± 0.018 | 0.4975 ± 0.018 | 0.4946 ± 0.016 | 0.4849 ± 0.013 | 0.4863 ± 0.013 |
| | KPCA | 0.8145 ± 0.018 | 0.7889 ± 0.017 | 0.6676 ± 0.057 | 0.6271 ± 0.023 | 0.7402 ± 0.029 | 0.6228 ± 0.021 | 0.6680 ± 0.045 | 0.6763 ± 0.026 | 0.4873 ± 0.057 |
| | PPCA | 0.6661 ± 0.019 | 0.6601 ± 0.018 | 0.6251 ± 0.030 | 0.6356 ± 0.021 | 0.6561 ± 0.026 | 0.6224 ± 0.016 | 0.6230 ± 0.010 | 0.6465 ± 0.037 | 0.4714 ± 0.043 |
| | MPK | 0.9336 ± 0.011 | 0.9321 ± 0.009 | 0.9163 ± 0.011 | 0.9059 ± 0.012 | 0.8989 ± 0.016 | 0.8991 ± 0.013 | 0.8852 ± 0.010 | 0.8825 ± 0.012 | 0.8324 ± 0.011 |
| MNAR | Mean | 0.8046 ± 0.013 | 0.8032 ± 0.014 | 0.7959 ± 0.015 | 0.7787 ± 0.013 | 0.7735 ± 0.014 | 0.7562 ± 0.014 | 0.7617 ± 0.015 | 0.7510 ± 0.022 | 0.7160 ± 0.012 |
| | MICE | 0.8046 ± 0.013 | 0.7959 ± 0.014 | 0.7710 ± 0.015 | 0.7456 ± 0.019 | 0.7384 ± 0.011 | 0.7057 ± 0.016 | 0.6692 ± 0.012 | 0.6699 ± 0.018 | 0.6790 ± 0.011 |
| | EM | 0.6684 ± 0.019 | 0.5971 ± 0.036 | 0.5190 ± 0.027 | 0.4787 ± 0.014 | 0.4639 ± 0.016 | 0.4560 ± 0.021 | 0.4500 ± 0.015 | 0.4330 ± 0.021 | 0.4226 ± 0.020 |
| | genRBF | 0.8668 ± 0.020 | 0.8654 ± 0.021 | 0.8674 ± 0.022 | 0.8691 ± 0.021 | 0.8669 ± 0.021 | 0.8574 ± 0.023 | 0.8455 ± 0.021 | 0.8218 ± 0.021 | 0.7837 ± 0.018 |
| | KPCA | 0.8195 ± 0.018 | 0.7881 ± 0.023 | 0.7770 ± 0.019 | 0.7512 ± 0.015 | 0.7424 ± 0.023 | 0.7146 ± 0.025 | 0.6972 ± 0.027 | 0.7051 ± 0.022 | 0.7697 ± 0.069 |
| | PPCA | 0.6608 ± 0.023 | 0.6275 ± 0.025 | 0.5630 ± 0.030 | 0.5366 ± 0.025 | 0.5135 ± 0.024 | 0.4938 ± 0.025 | 0.4638 ± 0.018 | 0.5422 ± 0.055 | 0.5421 ± 0.035 |
| | MPK | 0.9440 ± 0.009 | 0.9431 ± 0.010 | 0.9414 ± 0.010 | 0.9403 ± 0.012 | 0.9389 ± 0.011 | 0.9382 ± 0.009 | 0.9301 ± 0.014 | 0.9264 ± 0.014 | 0.9099 ± 0.010 |

Table 13: F1 Scores with corresponding standard deviations across various models and missing value proportions for the Spam dataset.

| Mechanism | Model | 0.05 | 0.1 | 0.2 | 0.3 | 0.4 | 0.5 | 0.6 | 0.7 | 0.8 |
|---|---|---|---|---|---|---|---|---|---|---|
| MCAR | Mean | 0.7886 ± 0.012 | 0.7706 ± 0.011 | 0.7576 ± 0.010 | 0.7552 ± 0.018 | 0.7296 ± 0.019 | 0.7129 ± 0.019 | 0.6636 ± 0.029 | 0.5836 ± 0.023 | 0.5188 ± 0.025 |
| | MICE | 0.7930 ± 0.013 | 0.7775 ± 0.013 | 0.7420 ± 0.010 | 0.7213 ± 0.015 | 0.7007 ± 0.023 | 0.6623 ± 0.018 | 0.6213 ± 0.020 | 0.6409 ± 0.019 | 0.5810 ± 0.021 |
| | EM | 0.6418 ± 0.013 | 0.6195 ± 0.015 | 0.5943 ± 0.026 | 0.5487 ± 0.025 | 0.4985 ± 0.038 | 0.4615 ± 0.038 | 0.4335 ± 0.031 | 0.4216 ± 0.021 | 0.4138 ± 0.030 |
| | genRBF | 0.8644 ± 0.021 | 0.8557 ± 0.023 | 0.8456 ± 0.022 | 0.8340 ± 0.025 | 0.8196 ± 0.030 | 0.7956 ± 0.028 | 0.7589 ± 0.018 | 0.7185 ± 0.017 | 0.6524 ± 0.013 |
| | KPCA | 0.7958 ± 0.017 | 0.7821 ± 0.025 | 0.6749 ± 0.047 | 0.6319 ± 0.053 | 0.5722 ± 0.024 | 0.5771 ± 0.056 | 0.5936 ± 0.057 | 0.6052 ± 0.074 | 0.6177 ± 0.067 |
| | PPCA | 0.6594 ± 0.018 | 0.6518 ± 0.017 | 0.6324 ± 0.018 | 0.6155 ± 0.030 | 0.6169 ± 0.031 | 0.6073 ± 0.032 | 0.5365 ± 0.058 | 0.5337 ± 0.022 | 0.4648 ± 0.024 |
| | MPK | 0.9339 ± 0.009 | 0.9222 ± 0.008 | 0.9080 ± 0.011 | 0.8935 ± 0.013 | 0.8750 ± 0.010 | 0.8622 ± 0.009 | 0.8401 ± 0.012 | 0.8067 ± 0.012 | 0.7639 ± 0.015 |
| MAR | Mean | 0.6648 ± 0.017 | 0.6352 ± 0.024 | 0.6160 ± 0.021 | 0.6052 ± 0.020 | 0.6715 ± 0.023 | 0.5282 ± 0.024 | 0.5195 ± 0.014 | 0.4788 ± 0.024 | 0.4602 ± 0.012 |
| | MICE | 0.6744 ± 0.017 | 0.6796 ± 0.023 | 0.6312 ± 0.031 | 0.6434 ± 0.023 | 0.6670 ± 0.025 | 0.6256 ± 0.017 | 0.6549 ± 0.010 | 0.6641 ± 0.036 | 0.4722 ± 0.037 |
| | EM | 0.6523 ± 0.019 | 0.6241 ± 0.017 | 0.5708 ± 0.033 | 0.5494 ± 0.024 | 0.6542 ± 0.016 | 0.4777 ± 0.046 | 0.4477 ± 0.025 | 0.4401 ± 0.029 | 0.4201 ± 0.058 |
| | genRBF | 0.5046 ± 0.012 | 0.5074 ± 0.009 | 0.5052 ± 0.011 | 0.5140 ± 0.007 | 0.5068 ± 0.018 | 0.4975 ± 0.018 | 0.4946 ± 0.016 | 0.4849 ± 0.013 | 0.4863 ± 0.013 |
| | KPCA | 0.8145 ± 0.018 | 0.7889 ± 0.017 | 0.6676 ± 0.057 | 0.6271 ± 0.023 | 0.7402 ± 0.029 | 0.6228 ± 0.021 | 0.6680 ± 0.045 | 0.6763 ± 0.026 | 0.4873 ± 0.057 |
| | PPCA | 0.6661 ± 0.019 | 0.6601 ± 0.018 | 0.6251 ± 0.030 | 0.6356 ± 0.021 | 0.6561 ± 0.026 | 0.6224 ± 0.016 | 0.6230 ± 0.010 | 0.6465 ± 0.037 | 0.4714 ± 0.043 |
| | MPK | 0.9336 ± 0.011 | 0.9321 ± 0.009 | 0.9163 ± 0.011 | 0.9059 ± 0.012 | 0.8989 ± 0.016 | 0.8991 ± 0.013 | 0.8852 ± 0.010 | 0.8825 ± 0.012 | 0.8324 ± 0.011 |
| MNAR | Mean | 0.8046 ± 0.013 | 0.8032 ± 0.014 | 0.7959 ± 0.015 | 0.7787 ± 0.013 | 0.7735 ± 0.014 | 0.7562 ± 0.014 | 0.7617 ± 0.015 | 0.7510 ± 0.022 | 0.7160 ± 0.012 |
| | MICE | 0.8046 ± 0.013 | 0.7959 ± 0.014 | 0.7710 ± 0.015 | 0.7456 ± 0.019 | 0.7384 ± 0.011 | 0.7057 ± 0.016 | 0.6692 ± 0.012 | 0.6699 ± 0.018 | 0.6790 ± 0.011 |
| | EM | 0.6684 ± 0.019 | 0.5971 ± 0.036 | 0.5190 ± 0.027 | 0.4787 ± 0.014 | 0.4639 ± 0.016 | 0.4560 ± 0.021 | 0.4500 ± 0.015 | 0.4330 ± 0.021 | 0.4226 ± 0.020 |
| | genRBF | 0.8668 ± 0.020 | 0.8654 ± 0.021 | 0.8674 ± 0.022 | 0.8691 ± 0.021 | 0.8669 ± 0.021 | 0.8574 ± 0.023 | 0.8455 ± 0.021 | 0.8218 ± 0.021 | 0.7837 ± 0.018 |
| | KPCA | 0.8195 ± 0.018 | 0.7881 ± 0.023 | 0.7770 ± 0.019 | 0.7512 ± 0.015 | 0.7424 ± 0.023 | 0.7146 ± 0.025 | 0.6972 ± 0.027 | 0.7051 ± 0.022 | 0.7697 ± 0.069 |
| | PPCA | 0.6608 ± 0.023 | 0.6275 ± 0.025 | 0.5630 ± 0.030 | 0.5366 ± 0.025 | 0.5135 ± 0.024 | 0.4938 ± 0.025 | 0.4638 ± 0.018 | 0.5422 ± 0.055 | 0.5421 ± 0.035 |
| | MPK | 0.9440 ± 0.009 | 0.9431 ± 0.010 | 0.9414 ± 0.010 | 0.9403 ± 0.012 | 0.9389 ± 0.011 | 0.9382 ± 0.009 | 0.9301 ± 0.014 | 0.9264 ± 0.014 | 0.9099 ± 0.010 |

Table 14: F1 Scores with corresponding standard deviations across various models and missing value proportions for the Student dataset.

| Mechanism | Model | 0.05 | 0.1 | 0.2 | 0.3 | 0.4 | 0.5 | 0.6 | 0.7 | 0.8 |
|---|---|---|---|---|---|---|---|---|---|---|
| MCAR | Mean | 0.8949 ± 0.010 | 0.8824 ± 0.010 | 0.8674 ± 0.013 | 0.8477 ± 0.010 | 0.8204 ± 0.009 | 0.7854 ± 0.009 | 0.7443 ± 0.011 | 0.6947 ± 0.016 | 0.6287 ± 0.008 |
| | MICE | 0.9109 ± 0.011 | 0.9118 ± 0.009 | 0.9065 ± 0.013 | 0.8908 ± 0.011 | 0.8904 ± 0.005 | 0.8555 ± 0.005 | 0.8347 ± 0.006 | 0.7943 ± 0.010 | 0.7120 ± 0.011 |
| | EM | 0.8800 ± 0.011 | 0.8639 ± 0.010 | 0.8190 ± 0.013 | 0.7753 ± 0.006 | 0.7352 ± 0.007 | 0.6807 ± 0.014 | 0.5959 ± 0.033 | 0.4298 ± 0.000 | 0.4298 ± 0.000 |
| | genRBF | 0.9823 ± 0.005 | 0.9761 ± 0.006 | 0.9545 ± 0.006 | 0.9379 ± 0.003 | 0.9096 ± 0.005 | 0.8789 ± 0.007 | 0.8345 ± 0.007 | 0.7779 ± 0.010 | 0.6925 ± 0.007 |
| | KPCA | 0.9229 ± 0.009 | 0.9198 ± 0.009 | 0.9111 ± 0.008 | 0.8908 ± 0.008 | 0.8850 ± 0.007 | 0.8594 ± 0.014 | 0.8404 ± 0.015 | 0.8008 ± 0.011 | 0.7185 ± 0.019 |
| | PPCA | 0.8931 ± 0.008 | 0.8791 ± 0.010 | 0.8455 ± 0.009 | 0.8059 ± 0.010 | 0.7726 ± 0.014 | 0.7353 ± 0.015 | 0.6960 ± 0.014 | 0.6454 ± 0.017 | 0.5825 ± 0.013 |
| | MPK | 0.9919 ± 0.003 | 0.9906 ± 0.002 | 0.9869 ± 0.002 | 0.9782 ± 0.004 | 0.9628 ± 0.006 | 0.9393 ± 0.005 | 0.9073 ± 0.008 | 0.8577 ± 0.015 | 0.7812 ± 0.012 |
| MAR | Mean | 0.9046 ± 0.010 | 0.8325 ± 0.014 | 0.8466 ± 0.013 | 0.8193 ± 0.012 | 0.8281 ± 0.006 | 0.9005 ± 0.012 | 0.6848 ± 0.028 | 0.6113 ± 0.015 | 0.5741 ± 0.014 |
| | MICE | 0.9074 ± 0.008 | 0.9020 ± 0.005 | 0.9004 ± 0.008 | 0.8815 ± 0.008 | 0.8954 ± 0.005 | 0.9032 ± 0.011 | 0.8127 ± 0.010 | 0.8363 ± 0.025 | 0.6251 ± 0.037 |
| | EM | 0.8858 ± 0.010 | 0.8435 ± 0.008 | 0.8216 ± 0.011 | 0.7577 ± 0.017 | 0.8050 ± 0.008 | 0.8945 ± 0.012 | 0.5937 ± 0.014 | 0.4298 ± 0.000 | 0.4298 ± 0.000 |
| | genRBF | 0.5043 ± 0.007 | 0.5060 ± 0.007 | 0.5040 ± 0.011 | 0.5053 ± 0.007 | 0.5079 ± 0.005 | 0.5052 ± 0.010 | 0.5028 ± 0.004 | 0.4981 ± 0.009 | 0.4937 ± 0.007 |
| | KPCA | 0.9239 ± 0.010 | 0.9217 ± 0.009 | 0.9147 ± 0.007 | 0.8936 ± 0.003 | 0.9059 ± 0.010 | 0.9089 ± 0.012 | 0.8515 ± 0.013 | 0.8621 ± 0.014 | 0.7762 ± 0.012 |
| | PPCA | 0.8983 ± 0.013 | 0.8821 ± 0.011 | 0.8402 ± 0.012 | 0.8557 ± 0.013 | 0.8241 ± 0.009 | 0.8947 ± 0.011 | 0.6870 ± 0.030 | 0.6165 ± 0.015 | 0.5725 ± 0.014 |
| | MPK | 0.9921 ± 0.002 | 0.9923 ± 0.002 | 0.9854 ± 0.003 | 0.9838 ± 0.004 | 0.9750 ± 0.006 | 0.9741 ± 0.005 | 0.9205 ± 0.007 | 0.9176 ± 0.008 | 0.8781 ± 0.019 |
| MNAR | Mean | 0.8855 ± 0.011 | 0.8905 ± 0.012 | 0.8889 ± 0.017 | 0.8379 ± 0.006 | 0.7590 ± 0.015 | 0.6411 ± 0.022 | 0.5788 ± 0.024 | 0.5334 ± 0.023 | 0.4978 ± 0.016 |
| | MICE | 0.8996 ± 0.015 | 0.8814 ± 0.012 | 0.8448 ± 0.017 | 0.8148 ± 0.013 | 0.7816 ± 0.015 | 0.6051 ± 0.021 | 0.5839 ± 0.020 | 0.5721 ± 0.033 | 0.5577 ± 0.025 |
| | EM | 0.8513 ± 0.010 | 0.7746 ± 0.016 | 0.5534 ± 0.025 | 0.4479 ± 0.010 | 0.4318 ± 0.004 | 0.4298 ± 0.000 | 0.4298 ± 0.000 | 0.4298 ± 0.000 | 0.4298 ± 0.000 |
| | genRBF | 0.9700 ± 0.003 | 0.9517 ± 0.004 | 0.9252 ± 0.006 | 0.8817 ± 0.011 | 0.8222 ± 0.018 | 0.7673 ± 0.020 | 0.7273 ± 0.019 | 0.6475 ± 0.023 | 0.5725 ± 0.025 |
| | KPCA | 0.9211 ± 0.008 | 0.9114 ± 0.018 | 0.8710 ± 0.013 | 0.8281 ± 0.020 | 0.7832 ± 0.022 | 0.6841 ± 0.008 | 0.7031 ± 0.018 | 0.7776 ± 0.038 | 0.7508 ± 0.055 |
| | PPCA | 0.9098 ± 0.010 | 0.9118 ± 0.010 | 0.8895 ± 0.013 | 0.8610 ± 0.015 | 0.8652 ± 0.014 | 0.8507 ± 0.010 | 0.8206 ± 0.013 | 0.7841 ± 0.014 | 0.7422 ± 0.014 |
| | MPK | 0.9904 ± 0.004 | 0.9867 ± 0.004 | 0.9721 ± 0.003 | 0.9487 ± 0.002 | 0.9042 ± 0.004 | 0.8618 ± 0.004 | 0.8077 ± 0.017 | 0.8507 ± 0.024 | 0.8779 ± 0.014 |

Table 15: F1 Scores with corresponding standard deviations across various models and missing value proportions for the Wine dataset.